\documentclass[10pt,journal,compsoc]{IEEEtran}
\usepackage{amsmath,amsfonts,amssymb}
\usepackage{algorithmic}
\usepackage{algorithm}
\usepackage{makecell}
\usepackage{array}
\usepackage[caption=false,font=normalsize,labelfont=sf,textfont=sf]{subfig}
\usepackage{textcomp}
\usepackage{stfloats}
\usepackage{url}
\usepackage{verbatim}
\usepackage{graphicx}
\usepackage{pifont}
\usepackage{tcolorbox}
\usepackage{enumitem}
\usepackage{cite}
\usepackage{multirow}
\usepackage{color}
\usepackage[colorlinks=true,linkcolor=blue, citecolor=blue, urlcolor=black]{hyperref}
\begin{document}

\title{Short-video Propagation Influence Rating: A New Real-world Dataset and A New Large Graph Model}

\author{Dizhan Xue, Shengsheng Qian, Chuanrui Hu, Changsheng Xu,~\IEEEmembership{Fellow,~IEEE}% <-this % stops a space

\thanks{Manuscript received MM DD, YY; revised MM DD, YY. \textit{(Corresponding author: Shengsheng Qian.)}}

\IEEEcompsocitemizethanks{
\IEEEcompsocthanksitem Dizhan Xue, Shengsheng Qian, and Changsheng Xu are with State Key Laboratory of Multimodal Artificial Intelligence Systems, Institute of Automation, Chinese Academy of Sciences, Beijing 100190, China, and also with School of Artificial Intelligence, University of Chinese Academy of Sciences (e-mail: xuedizhan17@mails.ucas.ac.cn; shengsheng.qian@nlpr.ia.ac.cn; csxu@nlpr.ia.ac.cn)
% \IEEEcompsocthanksitem 
% Jing Cui is with School of Computer Science and Engineering, Tianjin University of Technology, Tianjing 300384, China (e-mail: cjtxdy@stud.tjut.edu.cn).

\IEEEcompsocthanksitem 
Chuanrui Hu is with Qihoo 360 AI lab, Beijing 100026, China (e-mail: huchuanrui\_1206@163.com).
}% <-this % stops an unwanted space
}

\markboth{IEEE Transactions on Knowledge and Data Engineering}%
{Xue \MakeLowercase{\textit{et al.}}: Short-video Propagation Influence Rating: A New Real-world Dataset and A New Large Graph Model}

\IEEEtitleabstractindextext{
\begin{abstract}
  Short-video platforms have gained immense popularity, captivating the interest of millions, if not billions, of users globally.
  Recently, researchers have highlighted the significance of analyzing the propagation of short-videos, which typically involves discovering commercial values, public opinions, user behaviors, etc. 
  This paper proposes a new Short-video Propagation Influence Rating (SPIR) task and aims to promote SPIR from both the dataset and method perspectives.
  First, we propose a new Cross-platform Short-Video (XS-Video) dataset, which aims to provide a large-scale and real-world short-video propagation network across various platforms to facilitate the research on short-video propagation.
  Our XS-Video dataset includes 117,720 videos, 381,926 samples, and 535 topics across 5 biggest Chinese platforms, annotated with the propagation influence from level 0 to 9.
  To the best of our knowledge, this is the first large-scale short-video dataset that contains cross-platform data or provides all of the views, likes, shares, collects, fans, comments, and comment content.
  Second, we propose a Large Graph Model (LGM) named NetGPT, based on a novel three-stage training mechanism, to bridge heterogeneous graph-structured data with the powerful reasoning ability and knowledge of Large Language Models (LLMs).
  Our NetGPT can comprehend and analyze the short-video propagation graph, enabling it to predict the long-term propagation influence of short-videos.
  Comprehensive experimental results evaluated by both classification and regression metrics on our XS-Video dataset indicate the superiority of our method for SPIR.
  Our dataset and code will be open upon acceptance.
\end{abstract}

}

\maketitle

\IEEEdisplaynontitleabstractindextext

\IEEEpeerreviewmaketitle

\section{Introduction}
\IEEEPARstart{S}{hort-video} platforms, such as Tiktok (aka Douyin)\footnote{https://www.tiktok.com/}, Kwai (aka Kuaishou)\footnote{https://www.kwai.com/}, and Xigua\footnote{https://www.ixigua.com/}, have gained immense popularity, captivating the interest of millions, if not billions, of users globally.
The users are capable of publishing their activities, opinions, and thoughts by posting, commenting, and sharing various forms of online content, including text, images, and videos. 
Consequently, these overwhelming user behaviors and user-generated content have formed a huge and intricate propagation network, harboring a wealth of information about our society.
Recently, researchers have highlighted the significance of analyzing the propagation of short-videos \cite{tang2022knowledge,zhong2024predicting,song2024data}, which typically involves discovering commercial values, public opinions, user behaviors, etc.
Traditional research mainly focuses on forecasting the popularity of videos \cite{chen2016micro,xie2020multimodal,xie2021micro} measured by a single indicator, such as the number of views or likes. 
However, real-world propagation behaviors are more complex and can include sharing, collecting, commenting, etc.

\begin{figure}[t]
	\includegraphics[width=0.8\linewidth]{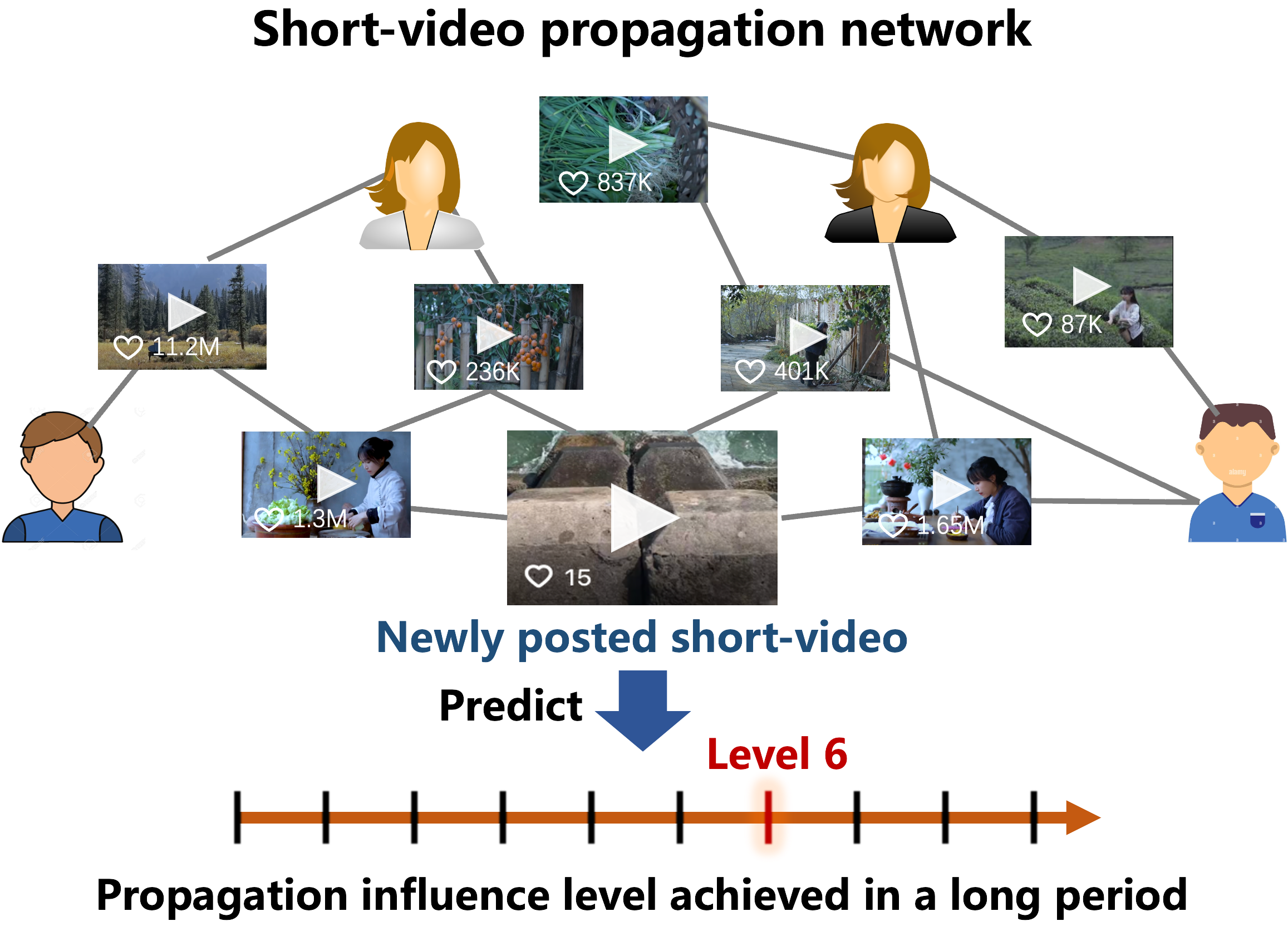}
		\centering
        %\vspace{-4mm}
	\caption{Short-video Propagation Influence Rating (SPIR): Predicting the influence level of a newly posted short-video that can be achieved in a long period.}
	\label{fig:task}
%\vspace{-4mm}
\end{figure}

\begin{table*}[t]
\centering
\footnotesize
\caption{A brief comparison of information collected in different short-video propagation datasets. \ding{51} means provided and \ding{55} means unprovided.}
\label{tab:dataset}
%\vspace{-2mm}
\setlength\tabcolsep{4pt}
\begin{tabular}{c|ccccccccccc}
\hline
Dataset                                         & \#Platforms             & Views                       & Likes                       & Shares                      & Collects                    & Fans                       & Comments                   & Comment Content & \#Videos & \#Samples & \#Users \\ \hline
Daum \cite{cha2009analyzing}       & 1 & \ding{51} & \ding{55} & \ding{55} & \ding{55} & \ding{55} & \ding{55} & \ding{55}                                & 196,037  & 196,037   & Unknown  \\
Youtube \cite{borghol2012untold}       & 1 & \ding{51} & \ding{51} & \ding{55} & \ding{55} & \ding{51} & \ding{51} & \ding{55}                                & 1,761  & 1,761   & Unknown  \\
NUS \cite{chen2016micro}       & 1 & \ding{51} & \ding{51} & \ding{55} & \ding{55} & \ding{51} & \ding{51} & \ding{55}                                & 303,242  & 303,242   & 98,166  \\
Xigua \cite{xie2020multimodal} & 1 & \ding{51} & \ding{55} & \ding{55} & \ding{55} & \ding{55} & \ding{55} & \ding{55}                                & 11,219   & 11,219    & 2,664   \\ 
MicroLens \cite{zhong2024predicting} & 1 & \ding{51} & \ding{51} & \ding{55} & \ding{55} & \ding{55} & \ding{55} & \ding{55}                                & 19,738   & 19,738    & 100,000 \\\hline
XS-Video (Ours)                                 & 5 & \ding{51} & \ding{51} & \ding{51} & \ding{51} & \ding{51} & \ding{51} & \ding{51}                                & 117,720  & 381,926   & 419,374        \\ \hline
\end{tabular}
%\vspace{-2mm}
\end{table*}

To address the above limitation, we attempt to evaluate the propagation influence by simultaneously considering as many interactive indicators as possible, including views, likes, shares, collects, comments, etc.
Moreover, as shown in Figure \ref{fig:task}, we propose a new task named \textit{Short-video Propagation Influence Rating} (SPIR), which aims to predict the long-term propagation influence of a newly posted short-video, under the background of a huge propagation network.
Different from traditional popularity prediction that forecasts a single interactive indicator over a short period (e.g., several hours \cite{xie2021micro,tang2022knowledge}), our SPIR predicts the comprehensive propagation influence assessed through multi-dimensional interactions accumulated over a long period.
SPIR requires a model to thoroughly analyze the multimodal content features, interactive information, and propagation relationships within a short-video.
Subsequently, a rating from 0 to 9 is assigned to represent the propagation influence of this video that can be achieved over a long period.
The development of SPIR systems can potentially benefit a wide spectrum of real-world applications on short-video platforms, including advertising \cite{yuan2022effect,xiao2023exploring}, recommendation \cite{cai2023two,zhang2024eeg}, and democracy \cite{chen2021positiveenergy,qin2023does}.
In this paper, we attempt to promote the research of SPIR from both the data and model perspectives, proposing the first cross-modal short-video propagation dataset and a novel large graph model for SPIR.

\begin{figure}[t]
	\includegraphics[width=1\linewidth]{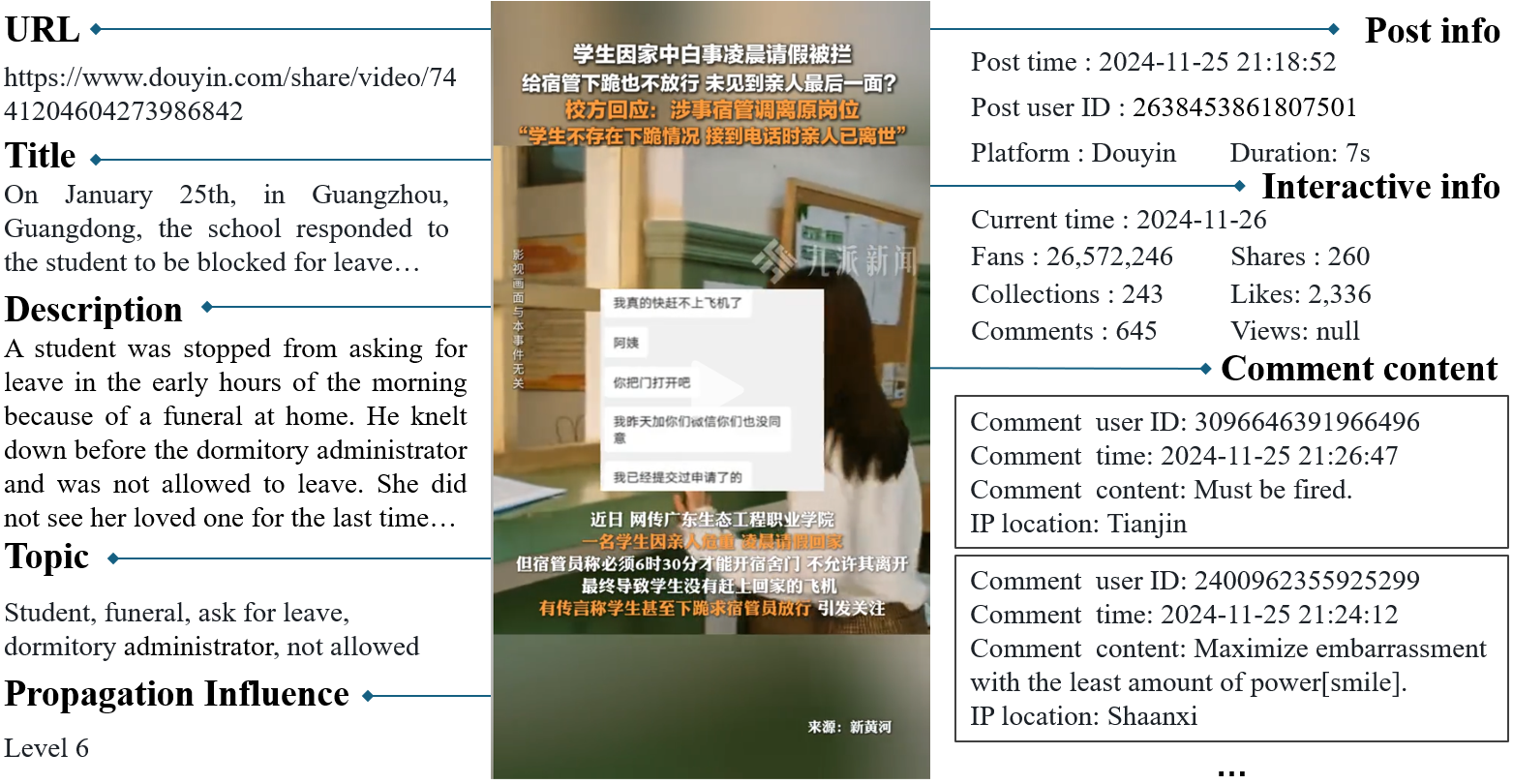}
		\centering
        %\vspace{-4mm}
	\caption{An example of short-video states/samples collected in our XS-Video dataset. The text is translated into English.}
	\label{fig:exp}
%\vspace{-4mm}
\end{figure}

To promote the research of SPIR and short-video propagation analysis, we propose a large-scale Cross-platform Short-Video (XS-Video) dataset, where videos are collected from the 5 biggest Chinese platforms (i.e., Douyin, Kuaishou, Xigua, Toutiao, and Bilibili).
Though hot topics and content are typically discussed and replicated across platforms, almost all existing datasets collect data from a single platform and lack a broad scope of short-videos.
As compared in Table \ref{tab:dataset}, our XS-Video dataset significantly surpasses the existing datasets in terms of platform numbers and interactive indicator coverage, while also containing comment content and more samples and users.
Specifically, we collect 381,926 states/samples of 117,720 short-videos within several days after their publication, belonging to 535 topics.
Different states/samples of the same video are distinguished by their sampling times, as well as the interactions and comment content at those times.
Furthermore, we annotate the propagation influence level of all short-videos according to their multi-dimensional interactive indicators over a two-week period following their publication, since almost all short-videos peak in interactions within two weeks.
We demonstrate a short-video sample in the XS-Video dataset in Figure \ref{fig:exp}.
Moreover, utilizing the entities and relations present in XS-Video, we construct a huge propagation graph consisting of 5.5 million nodes of videos, topics, comments, various interactions, etc., and 1.7 billion directed edges between these nodes.
More details about our dataset and the constructed propagation graph are provided in Section \ref{sec:graph}.
Compared to existing datasets, XS-Video provides more comprehensive information on video content, post, interactions, and comments, which is essential for the research on short-video propagation, including our SPIR task.

We build a comprehensive benchmark for SPIR by evaluating state-of-the-art methods, including Graph Neural Networks (GNNs), Large Language Models (LLMs), and multimodal LLMs. Surprisingly, these state-of-the-art methods are far from satisfactory for short-video propagation analysis. 
To improve the accuracy of SPIR and leverage the advancements in LLMs, we propose a Large Graph Model (LGM) named NetGPT. 
Our model can capture both the propagation network structure and heterogenous short-video features while utilizing the pretraining knowledge to predict the propagation influence level.
LLMs have recently shown powerful reasoning ability for text, images, and even videos \cite{liu2024deepseek,wu2024janus,xue2024few,wang2024qwen2,li2024llava}.
Current research has explored adapting LLMs for analyzing textual graph data \cite{tang2024graphgpt,li2024graph,wang2024can}.
However, bridging heterogeneous graph data and LLMs is still an open problem, especially as we cannot input our huge propagation graph into an LLM.
Our experiments also reveal that LLMs cannot directly learn the graph-structured relations in short-video propagation data, which results in poor performance for SPIR.
To address this limitation, we propose a three-stage training mechanism for establishing an effective LGM, including Heterogeneous Graph Pretraining, Supervised Language Fine-tuning, and Task-oriented Predictor Fine-tuning.
Finally, we can empower LLMs (e.g., Qwen2-VL \cite{wang2024qwen2} in our experiments) to comprehend and analyze the short-video propagation graph.
Our NetGPT model significantly outperforms state-of-the-art baselines, including Graph Neural Networks (GNNs), LLMs, and multimodal LLMs on our XS-Video dataset.

In brief, the contributions of our work are as follows:
\begin{itemize}[leftmargin=6mm]
    \item We propose the Short-video Propagation Influence Rating (SPIR) task, which aims to predict the long-term propagation influence of a newly posted short-video in a huge propagation network. Different from traditional popularity prediction that forecasts a single interactive indicator over a short period (e.g., several hours \cite{xie2021micro,tang2022knowledge}), SPIR predicts the comprehensive propagation influence assessed through multi-dimensional interactions over a long period.
    
    \item We propose XS-Video, a large-scale real-world dataset for short-video propagation comprising 381,926 samples with comprehensive content and interactive information collected from the 5 biggest Chinese platforms. Moreover, utilizing the entities and relations present in XS-Video, we construct a huge propagation graph consisting of 5.5 million nodes and 1.7 billion directed edges.

    \item We build a comprehensive benchmark for SPIR by evaluating state-of-the-art methods, including GNNs, LLMs, and multimodal LLMs. Surprisingly, these methods are far from satisfactory. These results motivate us to propose NetGPT, a novel Large Graph Model (LGM) that can bridge heterogeneous graph data and LLMs, based on a novel three-stage training mechanism. 
    %NetGPT can comprehend and analyze the short-video propagation network by effectively leveraging network structure, content features, and pretraining knowledge. 
    Experiments on XS-Video indicate that NetGPT significantly outperforms the state-of-the-art methods.

\end{itemize}

%\vspace{-1mm}
\section{Related Work}
\subsection{Short-Video Propagation Influence Analysis} 
Existing research on short-video propagation influence analysis mainly focuses on the popularity prediction of short-videos, such as predicting the views of newly posted videos \cite{roy2013towards,chen2016multi,jing2017low,tang2017popularity,ou2022mtaf}.
For example, Chen et al. \cite{chen2016micro} focus on the short-video representation and present a transductive multi-modal learning model for short-video popularity prediction.
Xie et al. \cite{xie2020multimodal} propose a multimodal variational encoder-decoder framework that considers the uncertain factors as the randomness for the mapping from the multimodal features to the popularity.
Xie et al. \cite{xie2021micro} propose a hierarchical multimodal variational encoder-decoder to predict the popularity of short-videos by comprehensively leveraging user information and the short-video content in a hierarchical
fashion.
Zhou et al. \cite{zhong2024predicting} present a multimodal retrieval-augmented popularity prediction model that improves prediction accuracy using relevant retrieved information.
However, the above research primarily focuses on forecasting some interactive indicators on a single platform, such as views or likes. While these indicators can provide a perspective into the popularity of videos, they fail to offer a comprehensive evaluation of video influence. In reality, the propagation behaviors are far more complex and multifaceted. Users engage in various actions that contribute to the cross-platform spread and influence of videos, including sharing, collecting, liking, and commenting videos. %Each of these behaviors represents a distinct dimension of user engagement and may significantly influence the video's diffusion and longevity within the network.
Therefore, to advance research in short-video propagation analysis, we propose a large-scale real-world XS-Video dataset. XS-Video collects cross-platform videos and offers a more comprehensive array of information, including videos, titles, topics, descriptions, 5 interactive indicators, and comment content, etc. Our dataset aims to facilitate a deeper research on propagation dynamics.

%\vspace{-1mm}
\subsection{Large Graph Model} Recent advancements in Large Language Models (LLMs) have showcased their impressive reasoning abilities not only with text but also with images and videos \cite{hu2024bliva,ge2024worldgpt,cha2024honeybee,li2025cumo}. This progress has led to an increasing focus on adapting LLMs for analyzing graph-structured data, especially textual graphs. 
For instance, Li et al. \cite{li2024graph} utilize LLM prompting on textual graphs, offering substantial potential to enhance graph transfer capabilities across diverse tasks and domains.
Tang et al. \cite{tang2024graphgpt} integrate LLMs with graph structural knowledge through graph instruction tuning, which includes a text-graph grounding component, a dual-stage instruction tuning approach, and a graph-text alignment projector.
Wang et al. \cite{wang2024can} propose Build-a-Graph Prompting and Algorithmic Prompting, two instruction-based approaches to enhance LLMs in solving natural language graph problems.

However, the above work focuses on textual graphs where the information of nodes and edges can transformed into text, thereby directly leveraging the textual reasoning ability of LLMs. These methods cannot be directly applied to our heterogeneous propagation graph of texts, videos, times, and scalars. Therefore, we propose NetGPT to capture both the propagation network structure and heterogenous short-video features, while utilizing the pretraining knowledge for SPIR.

\begin{figure*}[t]
	\includegraphics[width=0.80\linewidth]{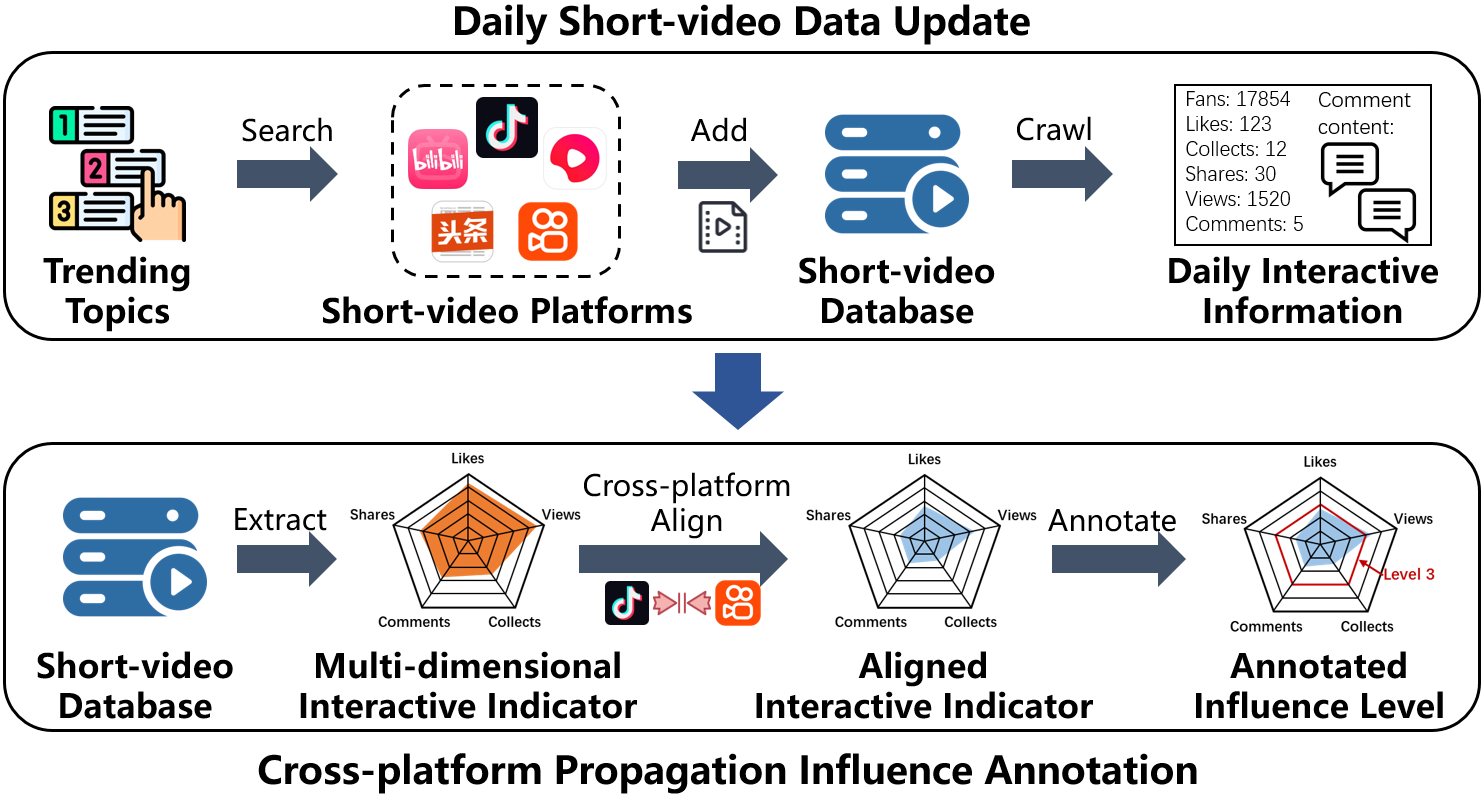}
		\centering
        %\vspace{-3mm}
	\caption{Brief construction procedure of our XS-Video dataset: (1) Daily update of new short-videos and the interactive information of already collected short-videos; (2) Alignment of multi-dimensional interactive indicators (collected 2 weeks later than the publication of videos) for annotating the video propagation influence levels.}
	\label{fig:framework1}
%\vspace{-4mm}
\end{figure*}

%\vspace{-1mm}
\section{Dataset}
\label{sec:data}
Analyzing propagation patterns on short-video platforms can aid in discovering high-value content, public opinions, and user preferences, thereby creating commercial and social value.
Our proposed dataset aims to provide a realistic and large-scale benchmark for the research on short-video propagation.
As briefly shown in Table \ref{tab:dataset} and Figure \ref{fig:exp}, compared to previous short-video propagation datasets \cite{cha2009analyzing,borghol2012untold,chen2016micro,xie2020multimodal,zhong2024predicting}, our dataset has three key advantages: (1) While existing datasets separately collect data from a single short-video platform, we collect short-videos from the 5 biggest Chinese platforms, which provide cross-platform propagation information; (2) Instead of focusing on a subset of interactive indicators, we gather a wide range of indicators such as views, likes, shares, collects, fans, and comments, which facilitate deep analysis of video interactions; (3) Our dataset contains more samples and users, while providing more comprehensive short-video content, including videos, titles, descriptions, topics, and the related comment content.
Figure \ref{fig:framework1} demonstrates a brief construction procedure of our dataset.

\subsection{Data Collection and Cleaning}
Utilizing the keywords of 535 trending topics on the Chinese Internet between November 25th, 2024 and January 2nd, 2025, we crawl site data of relevant short-videos with durations of less than 5 minutes from the 5 biggest Chinese platforms, i.e., Douyin, Kuaishou, Xigua, Toutiao, and Bilibili. 
Furthermore, we trace the interactive information of all videos until more than two weeks after their post.
Since almost all short-videos peak in interactive indicators and lose popularity within two weeks, our collected data is enough for assessing the long-term propagation influence of short-videos (as further analyzed in Appendix \ref{app:sta}).

Given active topics which include daily and historical trending topics on the Chinese Internet, we search and crawl all related short-videos posted in the last 30 days on 5 platforms.
However, some trending topics may be similar and correspond to the same concept.
Therefore, we conduct a daily data-cleaning procedure by manually merging similar topics, removing low-quality topics, and converting active topics without popularity to inactive topics.
For all collected short-videos, we crawl their interactive information everyday.
Since the interactive information of almost all videos peaks within 2 weeks, if the interactive information 2 weeks later than the post is crawled, we will stop crawling the video.
During constructing the data samples, we set the gap of current times of the same short-video as at least 2 days to avoid indistinguishable samples.
When constructing the data samples, we ensure that the current time gap between samples belonging to the same short-video is at least two days to prevent indistinguishable samples.
The whole data collection procedure is conducted from November 25th, 2024 to January 2nd, 2025.
Moreover, we replace user names with anonymous user IDs to avoid privacy problems.

%\vspace{-1mm}
\subsection{Cross-platform Propagation Influence Annotation}
\label{sec:cpia}
\textbf{Cross-platform indicator alignment.} To assess the long-term propagation influence of short-videos, we utilize the multi-dimensional interactive indicators accumulated in more than two weeks.
However, as the user amounts and activities significantly vary among different platforms (e.g., 0.6 billion daily active users of Douyin versus 0.2 billion monthly active users of Xigua), evaluation through raw interactive indicators cannot effectively annotate the influence of cross-platform short-videos.
Therefore, we leverage a novel cross-platform indicator alignment method. 
First, we additionally collect the short-videos posted by the 199 most popular influencers, covering 27 diverse domains such as social affairs, entertainment, dramas, shopping, celebrity, physical fitness, movie, gaming, relationship, beauty, technology, etc.
Since the same short-video posted by the same influencer on different platforms should roughly have a similar influence on these platforms, we compute the scaling factor $\alpha_*$ to minimize the Mean Squared Percentage Error (MSPE) of interactive indicators, where $*\in \{Douyin, Xigua, Toutiao, Bilibili\}$, for the platform $*$ as follows:
\begin{equation}
    \alpha_* = \mathop{\arg\min}_{\alpha_*} \frac{1}{|S_*|}\sum_{s\in S_*}\big(\frac{Ind_{Kuaishou}(s)-\alpha_*\cdot Ind_*(s)}{Ind_{Kuaishou}(s)+1}\big)^2,
\end{equation}
where $S_*$ are all short-videos posted on both Kuaishou and the platform $*$, and $Ind_*(s)$ is the interactive indicator (i.e., view, like, share, collect, or comments) of the short-video $s$ on the platform $*$.
Here, we choose Kuaishou as the central platform since it is the most popular among platforms that provide all 5 interactive indicators.

Finally, we design a stepped classification criteria to rate all short-videos in XS-Video into influence levels 0 to 9, using the aligned multi-dimensional indicators $\alpha_*\cdot Ind_*(v)$.
The criteria are designed to make sure that the influence levels follow a natural distribution \cite{rashid2005influence,cui2011should,liu2016you}.
The details of our classification criteria are explained in Appendix \ref{app:cri}.
%
%The statistics of the rated short-videos are shown in Figure \ref{}.

\begin{figure}[t]  
    \begin{minipage}{0.45\linewidth}
        \centerline{\includegraphics[width=\linewidth]{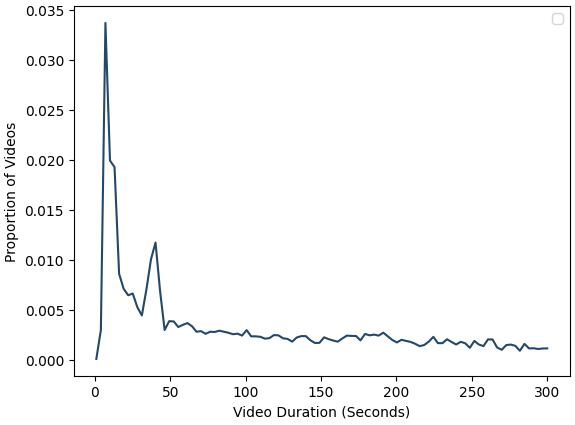}}
        %\vspace{-1mm}
        \centerline{\footnotesize{(a) Duration of short-videos}}
    \end{minipage}
    \hfill
    \begin{minipage}{0.45\linewidth}
        \centerline{\includegraphics[width=\linewidth]{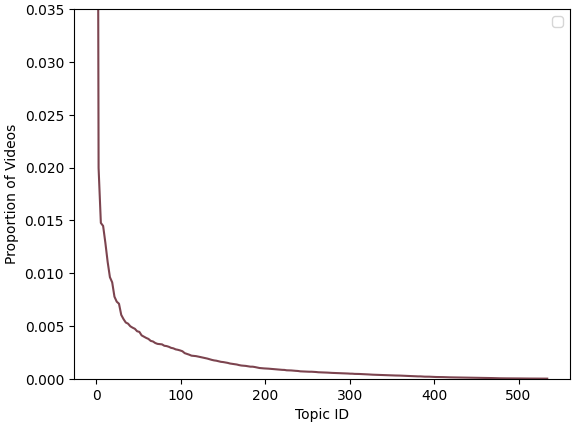}}
        %\vspace{-1mm}
        \centerline{\footnotesize{(b) Topic of short-videos}}
    \end{minipage}  
    \vfill
    \begin{minipage}{0.45\linewidth}
        \centerline{\includegraphics[width=\linewidth]{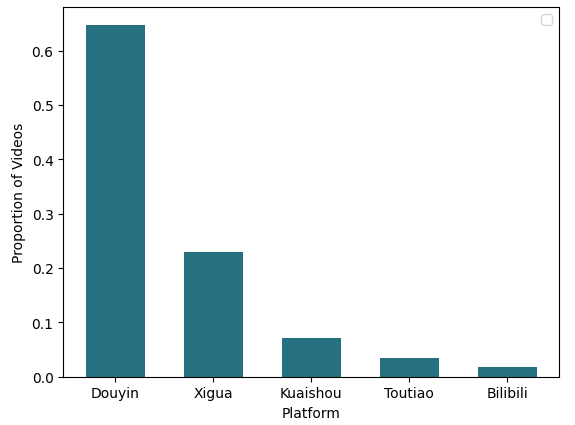}}
        %\vspace{-1mm}
        \centerline{\footnotesize{(c) Platform of short-videos}}
    \end{minipage}
    \hfill
    \begin{minipage}{0.45\linewidth}
        \centerline{\includegraphics[width=\linewidth]{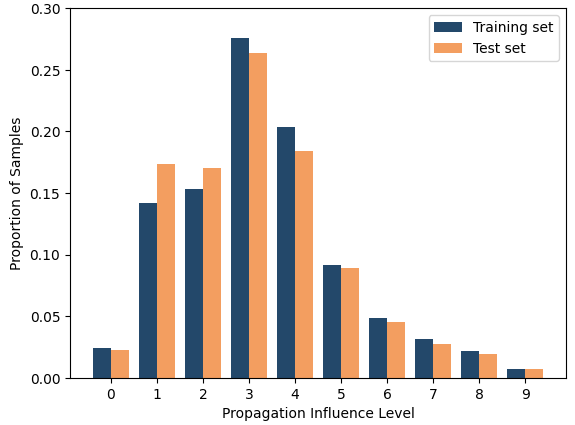}}
        %\vspace{-1mm}
        \centerline{\footnotesize{(d) Propagation influence levels}}
    \end{minipage}  
    \vfill
    \begin{minipage}{0.49\linewidth}
        \centerline{\includegraphics[width=\linewidth]{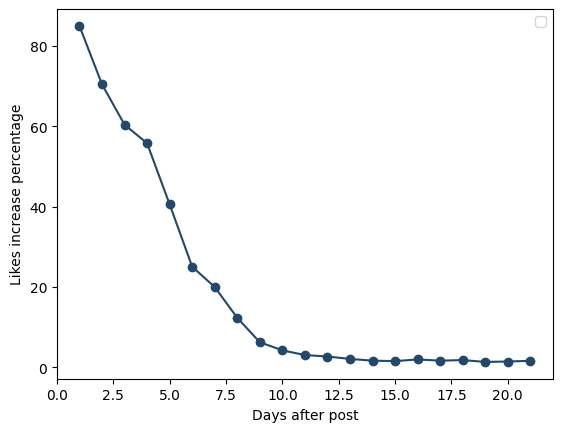}}
        %\vspace{-1mm}
        \centerline{\footnotesize{(e) Average increase percentage of likes}}
    \end{minipage}
    \hfill
    \begin{minipage}{0.49\linewidth}
        \centerline{\includegraphics[width=\linewidth]{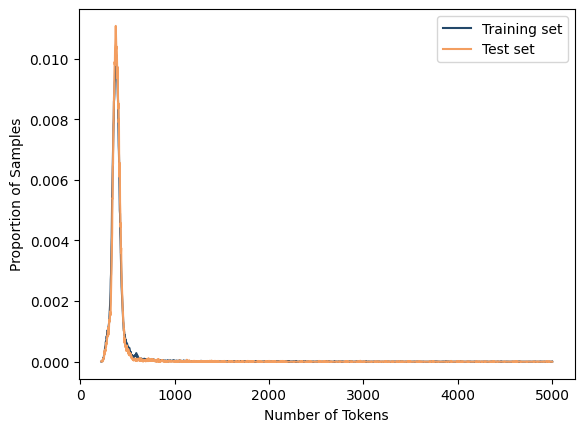}}
        %\vspace{-1mm}
        \centerline{\footnotesize{(f) Number of tokens}}
    \end{minipage}  
    \vfill
    \caption{Distributions of the short-videos and samples in our XS-Video dataset.}
    \label{fig:dis}
\end{figure}

\begin{table}[t]
\centering
\footnotesize
\caption{Criteria for Propagation Influence Level Annotation.}
\label{tab:level}
\setlength\tabcolsep{5pt}
\begin{tabular}{|c|c|c|c|c|c|}
\hline
 & Views     & Likes & Shares & Collects & Comments \\ \hline
Level 0 & 0         & 0     & 0      & 0        & 0        \\ \hline
Level 1 & $>$0         & $>$0     & $>$0      & $>$0        &  $>$0        \\ \hline
Level 2 & $>$100       & $>$10    & $>$10     & $>$10       & $>$10       \\ \hline
Level 3 & $>$100000    & $>$20    & $>$20     & $>$20       & $>$20       \\ \hline
Level 4 & $>$1000000   & $>$80    & $>$80     & $>$80       & $>$80       \\ \hline
Level 5 & $>$1000000   & $>$300   & $>$300    & $>$300      & $>$300      \\ \hline
Level 6 & $>$1000000   & $>$1000  & $>$1000   & $>$1000     & $>$1000     \\ \hline
Level 7 & $>$10000000  & $>$3000  & $>$3000   & $>$3000     & $>$3000     \\ \hline
Level 8 & $>$100000000 & $>$10000 & $>$10000  & $>$10000    & $>$10000    \\ \hline
Level 9 & $>$200000000 & $>$50000 & $>$50000  & $>$50000    & $>$50000    \\ \hline
\end{tabular}
\end{table}

\subsection{Criteria for Propagation Influence Level Annotation}
\label{app:cri}
To evaluate the propagation influence of short videos, our research adopts the cross-platform metrics alignment method described in \ref{sec:cpia} to systematically align the multidimensional interaction metrics accumulated over a two-week period. As shown in Table \ref{tab:level}, the criteria for dividing the propagation influence level are as follows: when all interactive indicators of a short-video are 0, its influence level is defined as level 0; for the division of the influence level from 1 to 9, a short video is determined to belong to this influence level when its interactive indicator on any of the five dimensions are  more than the corresponding level thresholds.
The criteria are designed to make sure that the influence levels follow a natural distribution \cite{rashid2005influence,cui2011should,liu2016you}, as shown in Figure \ref{fig:dis}(d).

\subsection{Data Analysis}
Our XS-Video dataset consists of 117,720 short-videos, with 304,362 training samples, and 77,564 test samples. 
The division between the training and test sets is determined by the post date cutoff of December 20th, 2024.
All training samples are posted on or before December 20th, 2024, while all test samples are posted after this date.
The ratio of the training and test data is around 4:1.

\textbf{Distribution of short-videos.} 
In Figure \ref{fig:dis} (a)-(c), we demonstrate the distributions of durations, topics, and platforms of the collected short-videos.
We can observe that the collected short-videos exhibit natural long-tail distributions.
While we collect videos with durations from 1s to 5min, the most popular durations are around 7s. Interestingly, 38s is also a subpeak.
For topics, we order topic IDs by their popularity.
We can find that a small number of head topics account for most of the popularity, while a large number of tail topics only have a small amount of heat.
For different platforms, such Matthew effect becomes more obvious.
While short-videos posted on Douyin, the most popular Chinese short-video platform, are more than half of all videos, smaller platforms occupy a small proportion of short-videos.
To sum up, long-tail distributions inherently exist in short-video data, which may increase the difficulty of analyzing short-video propagation.

\textbf{Distribution of samples.} 
In Figure \ref{fig:dis} (d), we demonstrate the distributions of the annotated propagation influence levels in the training and test sets.
The influence distributions are unimodal distributions, where level 3 is the peak of both sets.
When the influence further increases, the proportion of samples quickly decreases, exhibiting long-tail distributions on this half ($\geq$ level 3).
Moreover, since we adopt a practical division method of the training and test sets by a date boundary, the distributions of these two sets are not identical.
This slight mismatch of distributions also increases the difficulty of analyzing short-video propagation.

\label{app:sta}
\textbf{Increase process of interactive information.} To analyze the increase process of interactive information, we compute the daily likes increase percentage as follows:
\begin{equation}
    LIP(t) = E(\frac{Likes(t)-Likes(t-1)}{Likes(t)+1}),
\end{equation}
where $Likes(t)$ is the likes in the $t$-th day after the post of the short-video and $LIP(t)$ is the likes increase percentage of the $t$-th day.
In Figure \ref{fig:dis}(e), we plot the likes increase percentage of the $1$-st to $21$-th day after the post.
We can observe that the number of likes has become stable since the $10$-th day.
The similar trends can be observed for other interactive indicators.
These results verify our observation that the interactive information of almost all videos peaks within 2 weeks.
Therefore, we can utilize the interactive information later than 2 weeks after the post to annotate the final influence level achieved over a long period.

\textbf{Length of sample information.} We analyze the length of the sample informaiton in this section. We utilize the tokenizer of Qwen2-VL to tokenize the sample information in the json format and compute the number of tokens.
In Figure \ref{fig:dis}(f), we demonstrate the distribution of token lengths in XS-Video.
We can find that most samples have 200-500 tokens, while some samples with a large amount of comments have thousands of tokens.
Moreover, the distributions of token numbers in the training and test sets are similar.
Therefore, learning out-of-distribution patterns of long samples is a challenge in XS-Video.

\begin{figure}[h]  
    \begin{minipage}{0.95\linewidth}
        \centerline{\includegraphics[width=\linewidth]{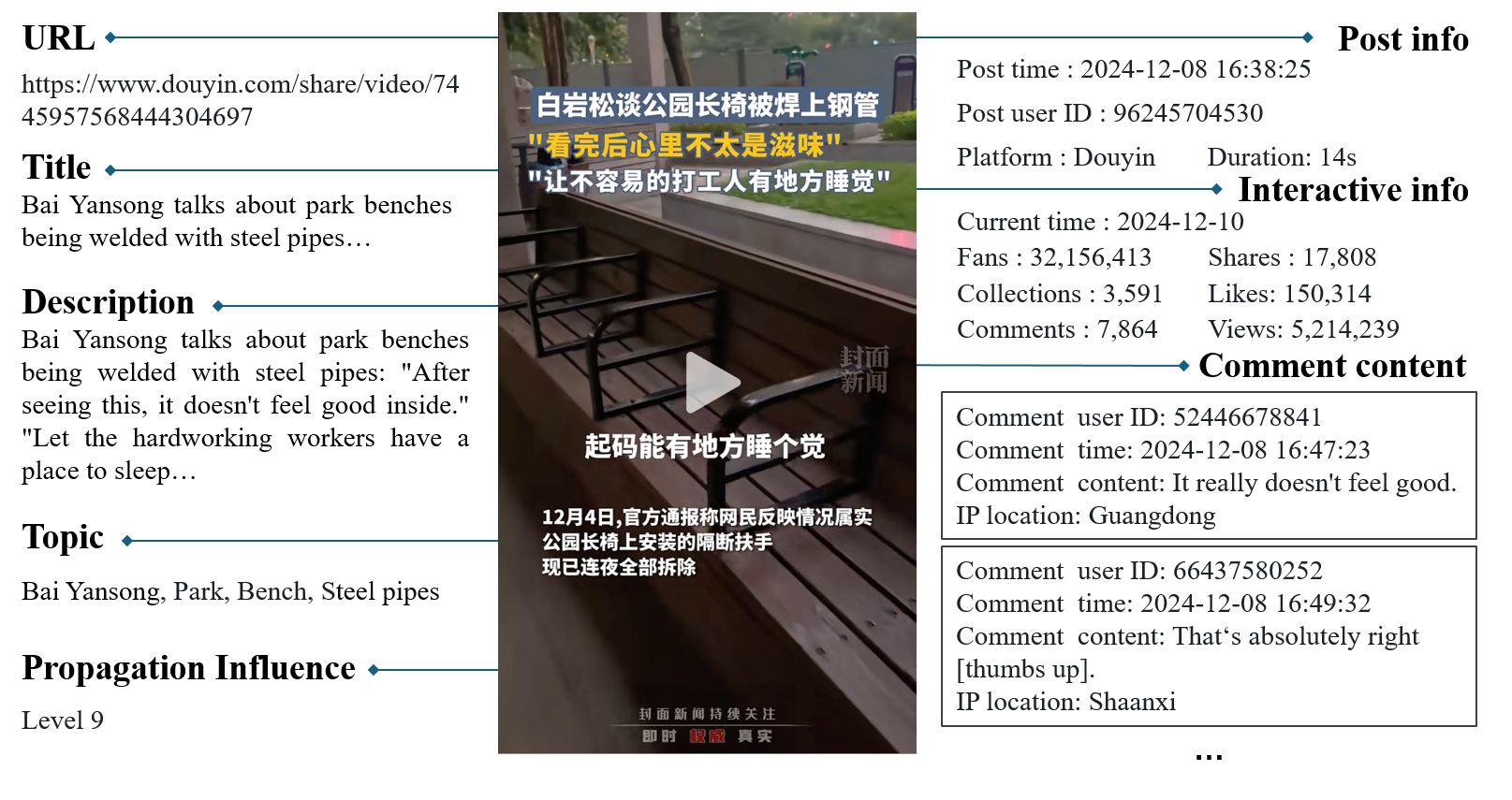}}
        %%\vspace{-1mm}
        \centerline{\footnotesize{(a)}}
    \end{minipage}
    \vfill
    %\vspace{2mm}
    \begin{minipage}{0.95\linewidth}
        \centerline{\includegraphics[width=\linewidth]{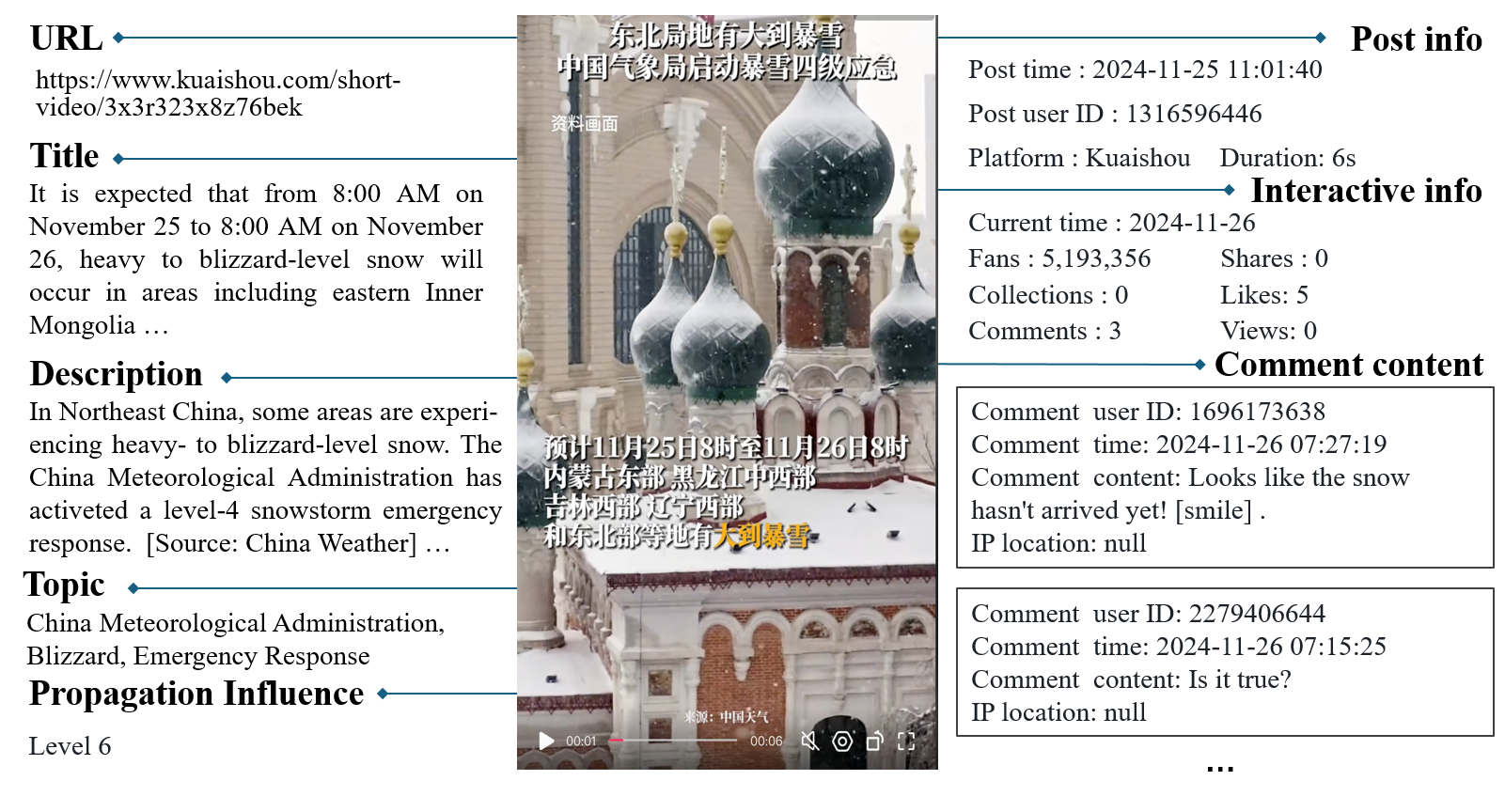}}
        %%\vspace{-1mm}
        \centerline{\footnotesize{(b)}}
    \end{minipage}  
    \vfill
    %\vspace{2mm}
    \begin{minipage}{0.95\linewidth}
        \centerline{\includegraphics[width=\linewidth]{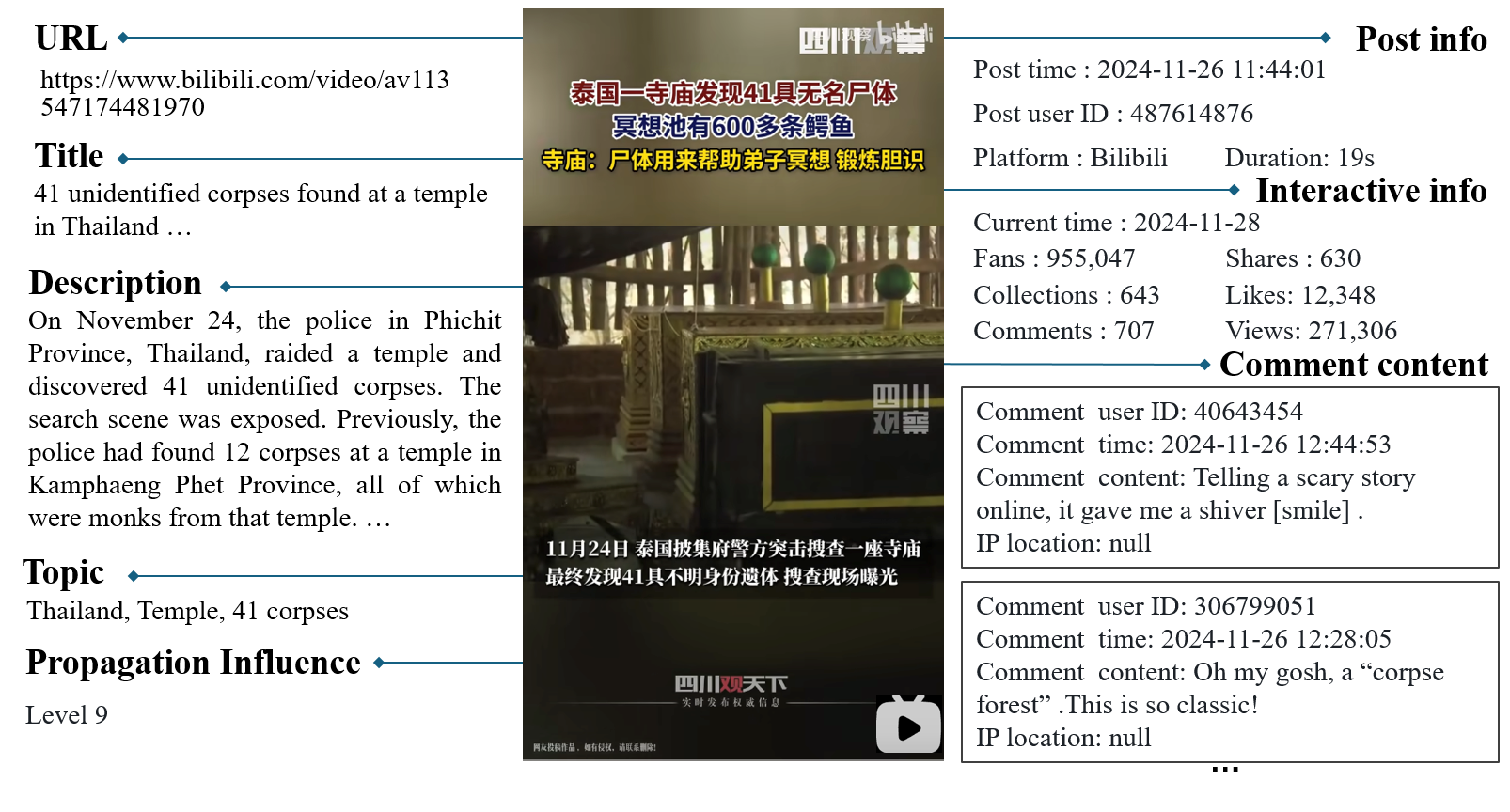}}
        %%\vspace{-1mm}
        \centerline{\footnotesize{(c)}}
    \end{minipage}  
    \vfill
    \caption{Examples of samples in our XS-Video dataset.}
    %%\vspace{-2mm}
    \label{fig:exp2}
\end{figure} 

\subsection{Example of Samples}
We exemplify more samples in XS-Video in Figure \ref{fig:exp2}.
In Figure \ref{fig:exp2}(a), the short-video has been posted for two days and gained a large amount of interactions. This short-video finally reaches the level-9 influence level in a long period.
In Figure \ref{fig:exp2}(b), the short-video has been posted for just 1 day and gained few interactions. However, this short-video will gain more interactions over a long period and finally reach the level-6 influence level.
In Figure \ref{fig:exp2}(c), the short-video has been posted for two days and gained a lot of interactions. Though the interactions of (c) is much less than those of (a), the user amount of Bilibili is much less than that of Douyin. After the cross-platform interactive information alignment, we annotate that this short-video finally reaches the level-9 influence level on its platform.

\begin{table*}[ht]
\centering
\footnotesize
\caption{Statistics of nodes in our constructed propagation graph of XS-Video.}
\label{tab:node}
\setlength\tabcolsep{3pt}
\begin{tabular}{|ccccc|}
\hline
\multicolumn{5}{|c|}{\textbf{Node}}                                                                                                                                                                                                                                                                                                              \\ \hline
\multicolumn{1}{|c|}{Name}        & \multicolumn{1}{c|}{Description}                                       & \multicolumn{1}{c|}{Format}    & \multicolumn{1}{c|}{Example}                                                                                                                                                  & Number    \\ \hline
\multicolumn{1}{|c|}{video}       & \multicolumn{1}{c|}{Visual content of a short-video}                   & \multicolumn{1}{c|}{MP4}       & \multicolumn{1}{c|}{-}                                                                                                                                                        & 381,926   \\ \hline
\multicolumn{1}{|c|}{platform}    & \multicolumn{1}{c|}{Post platform of a short-video}                    & \multicolumn{1}{c|}{Text}      & \multicolumn{1}{c|}{Douyin}                                                                                                                                                   & 5         \\ \hline
\multicolumn{1}{|c|}{topic}       & \multicolumn{1}{c|}{Topic keywords of a short-video}                   & \multicolumn{1}{c|}{Text}      & \multicolumn{1}{c|}{Student, funeral, ask for leave, dormitory administrator...}                                                                                              & 535       \\ \hline
\multicolumn{1}{|c|}{title}       & \multicolumn{1}{c|}{Title of a short-video}                            & \multicolumn{1}{c|}{Text}      & \multicolumn{1}{c|}{\begin{tabular}[c]{@{}c@{}}On January 25th, in Guangzhou, Guangdong, the school \\ responded to the student to be blocked for leave…\end{tabular}}        & 381,926   \\ \hline
\multicolumn{1}{|c|}{description} & \multicolumn{1}{c|}{Description of the video content}                  & \multicolumn{1}{c|}{Text}      & \multicolumn{1}{c|}{\begin{tabular}[c]{@{}c@{}}A student was stopped from asking for leave in the early \\ hours of the morning because of a funeral at home...\end{tabular}} & 381,926   \\ \hline
\multicolumn{1}{|c|}{time}        & \multicolumn{1}{c|}{Post time of a short-video}                        & \multicolumn{1}{c|}{Time}      & \multicolumn{1}{c|}{2024-11-25 21:18:52}                                                                                                                                      & 381,926   \\ \hline
\multicolumn{1}{|c|}{ctime}       & \multicolumn{1}{c|}{Sampling time of a short-video sample}              & \multicolumn{1}{c|}{Time}      & \multicolumn{1}{c|}{2024-11-26}                                                                                                                                               & 381,926   \\ \hline
\multicolumn{1}{|c|}{video\_time} & \multicolumn{1}{c|}{Duration of a short-video in seconds}              & \multicolumn{1}{c|}{Scalar}    & \multicolumn{1}{c|}{7}                                                                                                                                                        & 381,926   \\ \hline
\multicolumn{1}{|c|}{comment}     & \multicolumn{1}{c|}{Content of a comment of a short-video sample}      & \multicolumn{1}{c|}{Text+Time} & \multicolumn{1}{c|}{\begin{tabular}[c]{@{}c@{}}Maximize embarrassment with the least amount of power...\\ 2024-11-25 21:24:12\end{tabular}}                                   & 923,045   \\ \hline
\multicolumn{1}{|c|}{likes}       & \multicolumn{1}{c|}{Like number of a short-video sample}               & \multicolumn{1}{c|}{Scalar}    & \multicolumn{1}{c|}{2336}                                                                                                                                                     & 381,926   \\ \hline
\multicolumn{1}{|c|}{collects} & \multicolumn{1}{c|}{Collection number of a short-video sample}         & \multicolumn{1}{c|}{Scalar}    & \multicolumn{1}{c|}{243}                                                                                                                                                      & 381,926   \\ \hline
\multicolumn{1}{|c|}{views}       & \multicolumn{1}{c|}{View number of a short-video sample}          & \multicolumn{1}{c|}{Scalar}    & \multicolumn{1}{c|}{12563}                                                                                                                                                    & 381,926   \\ \hline
\multicolumn{1}{|c|}{shares}      & \multicolumn{1}{c|}{Share number of a short-video sample}              & \multicolumn{1}{c|}{Scalar}    & \multicolumn{1}{c|}{260}                                                                                                                                                      & 381,926   \\ \hline
\multicolumn{1}{|c|}{comments}    & \multicolumn{1}{c|}{Comment number of a short-video sample}            & \multicolumn{1}{c|}{Scalar}    & \multicolumn{1}{c|}{645}                                                                                                                                                      & 381,926   \\ \hline
\multicolumn{1}{|c|}{fans}        & \multicolumn{1}{c|}{Fans number of the poster of a short-video sample} & \multicolumn{1}{c|}{Scalar}    & \multicolumn{1}{c|}{26572246}                                                                                                                                                 & 381,926   \\ \hline
\multicolumn{1}{|c|}{All}         & \multicolumn{1}{c|}{Sum of above nodes}                                & \multicolumn{1}{c|}{-}         & \multicolumn{1}{c|}{-}                                                                                                                                                        & 5,506,697 \\ \hline
\end{tabular}
\end{table*}

\begin{table*}[ht]
\centering
\footnotesize
\caption{Statistics of edges in our constructed propagation graph of XS-Video.}
\label{tab:edge}
\begin{tabular}{|ccccc|}
\hline
\multicolumn{5}{|c|}{\textbf{Edge}}                                                                                                                                                                                                    \\ \hline
\multicolumn{1}{|c|}{Head Entity} & \multicolumn{1}{c|}{Relation}               & \multicolumn{1}{c|}{Tail Entity} & \multicolumn{1}{c|}{Description}                                                                  & Number        \\ \hline
\multicolumn{1}{|c|}{platform}    & \multicolumn{1}{c|}{is\_platform\_of}       & \multicolumn{1}{c|}{video}       & \multicolumn{1}{c|}{The platform of a short-video}                                                & 381,926       \\ \hline
\multicolumn{1}{|c|}{topic}       & \multicolumn{1}{c|}{is\_topic\_of}          & \multicolumn{1}{c|}{video}       & \multicolumn{1}{c|}{The topic of a short-video}                                                   & 381,926       \\ \hline
\multicolumn{1}{|c|}{title}       & \multicolumn{1}{c|}{is\_title\_of}          & \multicolumn{1}{c|}{video}       & \multicolumn{1}{c|}{The title of a short-video}                                                   & 381,926       \\ \hline
\multicolumn{1}{|c|}{description} & \multicolumn{1}{c|}{is\_description\_of}    & \multicolumn{1}{c|}{video}       & \multicolumn{1}{c|}{The description of a short-video}                                             & 381,926       \\ \hline
\multicolumn{1}{|c|}{time}        & \multicolumn{1}{c|}{is\_post\_time\_of}     & \multicolumn{1}{c|}{video}       & \multicolumn{1}{c|}{The post time of a short-video}                                               & 381,926       \\ \hline
\multicolumn{1}{|c|}{ctime}       & \multicolumn{1}{c|}{is\_current\_time\_of}  & \multicolumn{1}{c|}{video}       & \multicolumn{1}{c|}{The current time of a short-video sample}                                     & 381,926       \\ \hline
\multicolumn{1}{|c|}{video\_time} & \multicolumn{1}{c|}{is\_duration\_time\_of} & \multicolumn{1}{c|}{video}       & \multicolumn{1}{c|}{The duration time of a short-video}                                           & 381,926       \\ \hline
\multicolumn{1}{|c|}{comment}     & \multicolumn{1}{c|}{is\_comment\_of}        & \multicolumn{1}{c|}{video}       & \multicolumn{1}{c|}{A comment of a short-video sample}                                            & 923,045       \\ \hline
\multicolumn{1}{|c|}{likes}       & \multicolumn{1}{c|}{is\_likes\_of}          & \multicolumn{1}{c|}{video}       & \multicolumn{1}{c|}{The like number of a short-video sample}                                      & 381,926       \\ \hline
\multicolumn{1}{|c|}{collects} & \multicolumn{1}{c|}{is\_collects\_of}    & \multicolumn{1}{c|}{video}       & \multicolumn{1}{c|}{The collection number of a short-video sample}                                & 381,926       \\ \hline
\multicolumn{1}{|c|}{views}       & \multicolumn{1}{c|}{is\_views\_of}          & \multicolumn{1}{c|}{video}       & \multicolumn{1}{c|}{The play/view number of a short-video sample}                                 & 381,926       \\ \hline
\multicolumn{1}{|c|}{shares}      & \multicolumn{1}{c|}{is\_shares\_of}         & \multicolumn{1}{c|}{video}       & \multicolumn{1}{c|}{The share number of a short-video sample}                                     & 381,926       \\ \hline
\multicolumn{1}{|c|}{comments}    & \multicolumn{1}{c|}{is\_comments\_of}       & \multicolumn{1}{c|}{video}       & \multicolumn{1}{c|}{The comment number of a short-video sample}                                   & 381,926       \\ \hline
\multicolumn{1}{|c|}{fans}        & \multicolumn{1}{c|}{is\_fans\_of}           & \multicolumn{1}{c|}{video}       & \multicolumn{1}{c|}{The fans number of the poster of a short-video sample}                        & 381,926       \\ \hline
\multicolumn{1}{|c|}{video}       & \multicolumn{1}{c|}{has\_same\_author\_as}  & \multicolumn{1}{c|}{video}       & \multicolumn{1}{c|}{The head (posted earlier) has the same author as the tail (posted later)}     & 5,372,152     \\ \hline
\multicolumn{1}{|c|}{video}       & \multicolumn{1}{c|}{has\_same\_topic\_as}   & \multicolumn{1}{c|}{video}       & \multicolumn{1}{c|}{The head (posted earlier) has the same topic as the tail (posted later)}      & 1,655,971,954 \\ \hline
\multicolumn{1}{|c|}{video}       & \multicolumn{1}{c|}{is\_history\_of}        & \multicolumn{1}{c|}{video}       & \multicolumn{1}{c|}{The head (posted earlier) is the historical state of the tail (posted later)} & 484,364       \\ \hline
\multicolumn{1}{|c|}{-}         & \multicolumn{1}{c|}{-}     & \multicolumn{1}{c|}{-}           & \multicolumn{1}{c|}{Sum of above edges}                                                                            & 1,667,716,553     \\ \hline
\end{tabular}
\end{table*}

\subsection{Short-video Propagation Graph}
\label{sec:graph}
Utilizing the entities and relations present in XS-Video, we construct a huge propagation graph $G=\{V, E, S\}$ with the label set $Y$.
Here $V$ is the vertex set, containing 5 million nodes of videos, platforms, topics, titles, descriptions, comments, various interactive indicators, etc.
$E$ is the edge set, containing 1.7 billion directed edges between nodes, which represent the belonging relations between videos and their attributions, common author relations between videos, common topic relations between videos, chronological relations between samples of the same videos, etc.
Specially, $S\subset V$ is the set of all video nodes and $Y$ is the set of annotated propagation influence levels of all videos.

Utilizing the entities and relations present in XS-Video, we construct a huge propagation graph consisting of 5.5 million nodes of videos, topics, comments, various interactions, etc., and 1.7 billion directed edges between these nodes.
The information about nodes and edges in our graph is shown in Tables \ref{tab:node} and \ref{tab:edge}, respectively.
The short-video propagation graph is a typical heterogenous graph with both heterogenous nodes and edges.
In brief, the nodes include the video nodes, the static attribution nodes of videos, and the dynamic interactive nodes of videos.
The edges include the edges from the attribution nodes to videos, the interactive nodes to videos, and the video nodes to video nodes.

\begin{figure*}[ht]
	\includegraphics[width=0.80\linewidth]{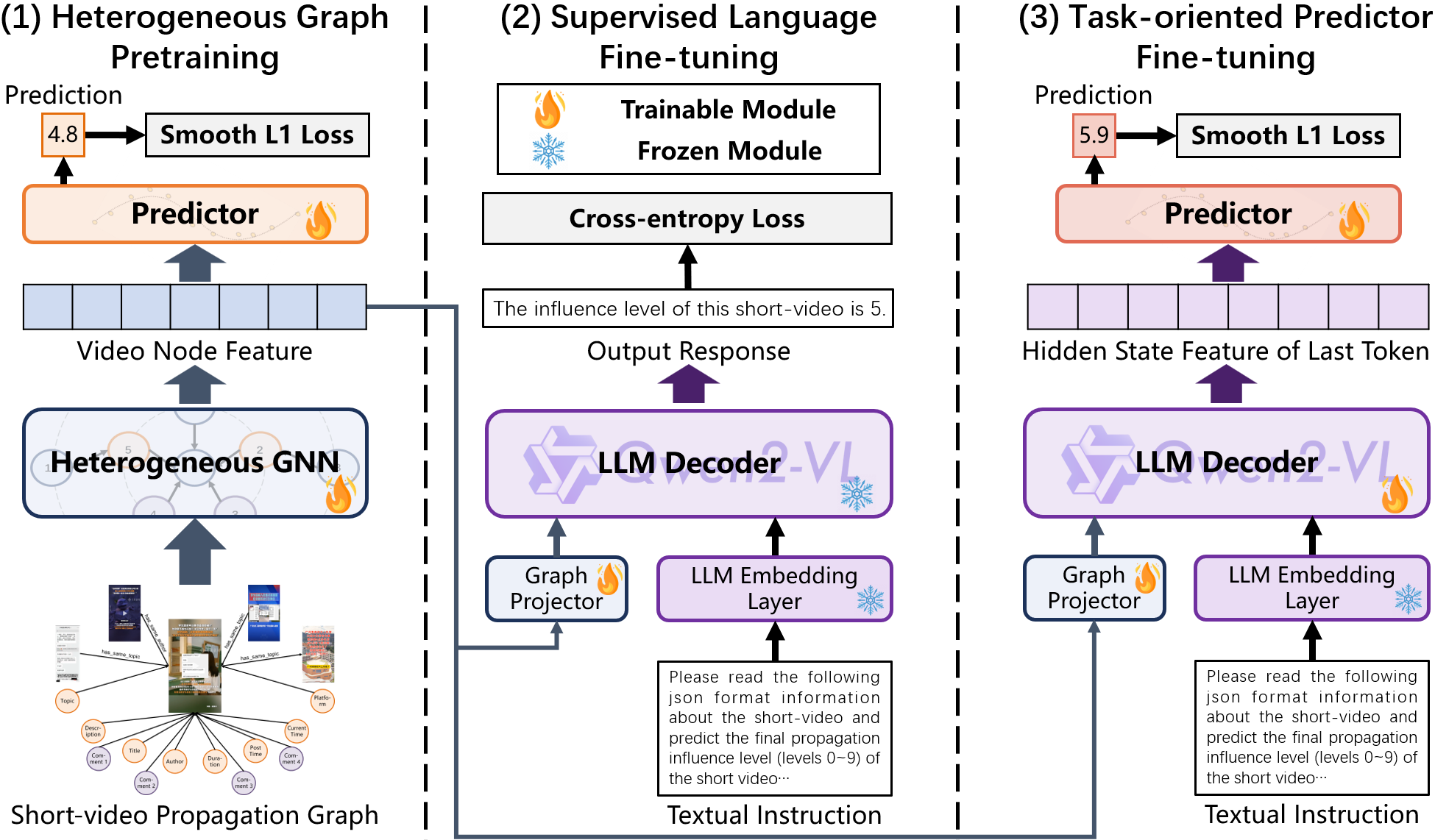}
		\centering
  %\vspace{-2mm}
	\caption{Framework of our proposed NetGPT model: (1) Pretrain a heterogeneous GNN to obtain the features of the video nodes; (2) Train a graph projector to bridge GNN feature space and the LLM embedding space by supervised instruction fine-tuning; (3) Fine-tuning the model with an additional predictor to obtain the final influence level of the short-videos.}
	\label{fig:framework2}
%\vspace{-4mm}
\end{figure*}

\section{Method}
Short-video propagation network is typical graph-structured data, where edges (such as video-author and video-video) represent complex propagation relations between entries.
State-of-the-art methods for analyzing graph-structured data usually adopt Graph Neural Networks (GNNs) \cite{hong2020attention,yussif2023self,hu2024efficient}.
Since Large Language Models (LLMs) have recently shown powerful reasoning ability for textual data, researchers have attempted to adapt LLMs to understand graph-structured data \cite{tang2024graphgpt,li2024graph,wang2024can}.
However, existing research focuses on textual graphs where the information of nodes and edges can be transformed into text, thereby directly leveraging the textual reasoning ability of LLMs.
These methods cannot be directly applied to our heterogeneous propagation graph of short-videos.
To analyze the short-video propagation graph, where nodes can be texts, times, scalars, and videos, we propose a novel Large Graph Model (LGM) named NetGPT
NetGPT fuses open-source LLM (i.e., Qwen2-VL \cite{wang2024qwen2} in our experiments) and heterogeneous GNN to effectively comprehend and analyze the short-video propagation graph.
Figure \ref{fig:framework2} illustrates the framework of our three-stage method.

%\vspace{-1mm}
\subsection{Heterogeneous Graph Pretraining} 
\label{sec:hgp}
Given the short-video propagation graph $G=\{V, E, S\}$ formulated in Section \ref{sec:graph} and Tables \ref{tab:node}-\ref{tab:edge}, we first extract the raw features of all heterogenous nodes in $V$.
For each node $v$ of the video modality, we utilize the ViT \cite{dosovitskiy2020image} in Qwen2-VL-7B-Instruct\footnote{https://huggingface.co/Qwen/Qwen2-VL-7B-Instruct} to extract the video feature with an average pooling operation, as follows:
\begin{equation}
\label{eq:vf}
    f^v = Avg{\rm{-}}Pooling(ViT(v))\in \mathbb{R}^{3584},
\end{equation}
where $Avg{\rm{-}}Pooling(\cdot)$ is the average pooling operation that computes the average feature of the spatial-temporal features outputted by the ViT.
For each node $v$ of the text modality (including platforms, topics, titles, and descriptions), we utilize the RoBERTa \cite{Liu2019RoBERTaAR,cui2021pre} of the Chinese-RoBERTa-wwm-ext-large version\footnote{https://huggingface.co/hfl/chinese-roberta-wwm-ext-large} to extract the text feature, as follows:
\begin{equation}
    f^v = RoBERTa(v)\in \mathbb{R}^{1024}.
\end{equation}
For each node $v$ of the time modality (including times and ctimes), as $v$ is formulated in the timestamp format, we utilize Sinusoidal Encoding \cite{NIPS2017_3f5ee243} to compute the time feature, as follows:
\begin{equation}
\begin{aligned}
    f^v_{2i} = sin(v/10000^{2i/512}), i=0,...,255,\\
    f^v_{2i+1} = cos(v/10000^{2i/512}), i=0,...,255,
\end{aligned}
\end{equation}
where $f^v\in \mathbb{R}^{512}$ can capture the relative time-spanning information.
For each node $v$ of the scalar value (including likes, collects, views, shares, comments, and fans), we directly use the logarithmic values as the raw feature, as follows:
\begin{equation}
    f^v = log(v+1)\in \mathbb{R}^{1}.
\end{equation}
Specially, each comment node contains the content text and the comment time, thereby we separately extract two features and concatenate them to the comment feature, as follows:
\begin{equation}
    f^v = (f^v_{text}||f^v_{time})\in \mathbb{R}^{1536},
\end{equation}
where $(\cdot||\cdot)$ is the vector concatenation operation, $f^v_{text}\in\mathbb{R}^{1024}$ is the text feature extracted by RoBERTa and $f^v_{time}\in\mathbb{R}^{512}$ is the time feature encoded by Sinusoidal Encoding.
After obtaining the raw features $F=\{f_v|v\in V\}$ of all nodes, we compute the high-level features that capture both node information and edge structures by adopting heterogenous GNN, as follows:
\begin{equation}
\label{eq:rgcn}
    F' = \{f'_v|v\in V\} = GNN(F_{raw}, E),
\end{equation}
where $f'_v\in \mathbb{R}^{d_g}$ is the output feature of the node $v$, $d_g$ is the dimension of the feature, and $GNN(\cdot,\cdot)$ is a 2-layer RGCN \cite{schlichtkrull2018modeling}.
Then, we add a 1-layer predictor layer and pretrain the GNN with the predictor by Smooth L1 loss \cite{girshick2015fastrcnn}, as follows:
\begin{equation}
\label{eq:loss1}
\begin{aligned}
    \hat{y}_v&=9\cdot sigmoid(\boldsymbol{W}_1f'_v+b_1)\in (0,9),\\
    \mathcal{L}_{pt} &= \frac{1}{|S_{tra}|}\sum_{v\in S_{tra}} SmoothL\textit{1}\big(\hat{y}_v, y_v\big),\\%\ell\big(\hat{y}_v, y_v\big),\\
 %   \ell\big(\hat{y}_v, y_v\big) &= \begin{cases}
	% 0.5\cdot (\hat{y}_v - y_v)^2, &if |\hat{y}_v - y_v|<1\\
	% |\hat{y}_v - y_v|-0.5, &otherwise	
	% 	   \end{cases}
\end{aligned}
\end{equation}
where $W_1\in\mathbb{R}^{1\times d_g}$ and $b_1\in \mathbb{R}$ are trained parameters of the predictor layer, $y_v$ is the ground truth influence level, $sigmoid(\cdot)$ is the sigmoid function that scales output into $(0,1)$, $S_{tra}$ is the set of all video nodes in the training set, and $SmoothL\textit{1}(\cdot, \cdot)$ is the Smooth L1 loss.
After optimizing the pretraining loss $\mathcal{L}_{pt}$, we obtain the heterogeneous graph encoder $GNN(\cdot, \cdot)$, which will be adopted in the following stages.

\subsection{Supervised Language Fine-tuning}
\label{sec:slf}
Given the pretrained graph encoder $GNN(\cdot, \cdot)$ and an open-source LLM $LLM(\cdot)$ (e.g., Qwen2-VL in our experiments), this stage aims to align the graph feature space of GNN and the token embedding space of LLM.
Frozen the pretrained GNN, we utilize a linear projector to transform $f'_v$ into the token embedding space, as follows:
\begin{equation}
\label{eq:pro}
    e_v = \boldsymbol{W}_2 f'_v + b_2,
\end{equation}
where $\boldsymbol{W}_2\in\mathbb{R}^{d_{lm}\times d_g}$ and $b_2\in\mathbb{R}^{d_{lm}}$ are trainable parameters, and $d_{lm}$ is the token embedding dimension of the LLM.
Next, we construct the input instruction $ins$ for LLM, which is ``\textit{You are a helpful assistant. Graph: $(v,E)$. Please read the following json format information about the short-video and predict the final propagation influence level (levels 0~9) of the short video: $x_v$}", where $x_v$ is the textual information of the sample (as shown in Figure \ref{fig:exp} except the influence level) in the json format. More details of our instruction are introduced in Appendix \ref{app:prompt}.
Then, we obtain the input token embeddings of $ins$, where the embedding of $(v,E)$ is $e_v$ computed by Equations \ref{eq:rgcn} and \ref{eq:pro}, and the embeddings of other textual tokens are obtained by the embedding layer of the LLM.
The ground truth output of this stage is constructed as response $res$=``\textit{The influence level of this short-video is $y_v$}", where $y_v$ is the ground truth influence level of $v$.
The optimization objective of this stage is maximizing the generation probability of $res$ with the input $ins$, as follows:
\begin{equation}
\label{eq:loss2}
    \max P_{LLM}(res|ins),
\end{equation}
where $P_{LLM}(\cdot|\cdot)$ is the generation probability function of the LLM.
Following the original implementation of Qwen2-VL \cite{wang2024qwen2}, we leverage the cross-entropy loss to optimize the Equation \ref{eq:loss2}.
Specially, to preserve the pretrained knowledge and avoid the model corruption introduced by fusing GNN and LLM, we only train the parameters $\boldsymbol{W}_2$ and $b_2$ of the graph feature projector, while keeping all parameters of the pretrained GNN and LLM frozen.

\begin{figure}[h]
	\includegraphics[width=0.98\linewidth]{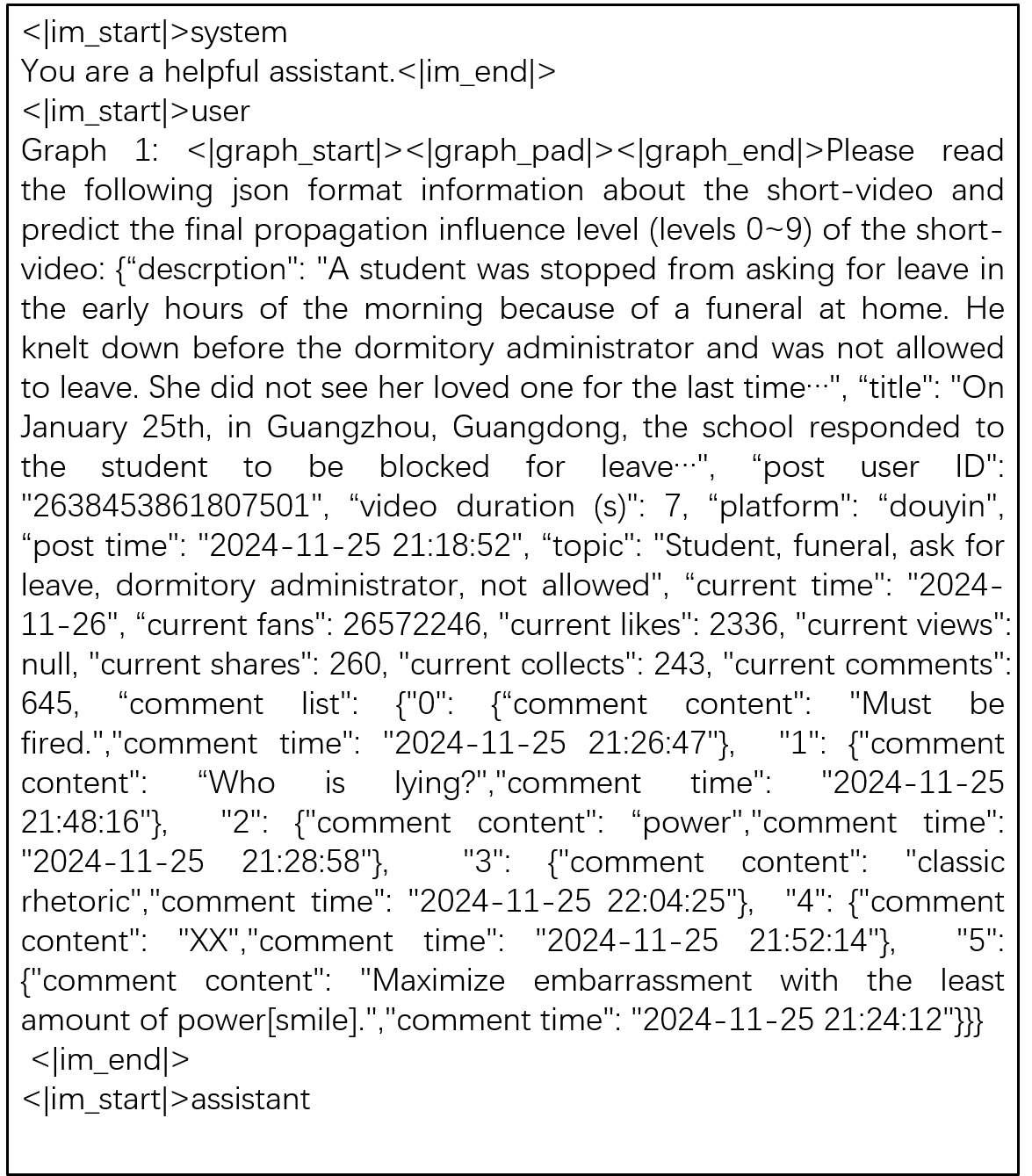}
		\centering
        %%\vspace{-2mm}
	\caption{An example of the input instruction in our method. The text is translated into English.}
	\label{fig:prompt}
\end{figure}

\subsection{Prompt Instruction for Fine-tuning}
\label{app:prompt}
In supervised language fine-tuning and the following task-oriented predictor fine-tuning, we combine the graph feature and the textual information for SPIR.
Assume $x_v$ is the textual information of the video $v$ in the json format, the detailed input prompt is as follows:
\begin{equation}
\begin{aligned}
&<|im\_start|>system\\
&You\ are\ a\ helpful\ assistant.<|im\_end|>\\
&<|im\_start|>user\\
&Graph\ 1:\ <|graph\_start|><|graph\_pad|>\\
&<|graph\_end|> Please\ read\ the\ following\ json\ format\\
&information\ about\ the\ short{\rm{-}}video\ and\ predict\ the\ final\\ 
&propagation\ influence\ level\ (levels\ 0~9)\ of\ the\ short{\rm{-}}video:\\
&x_v <|im\_end|>\\
&<|im\_start|>assistant
\end{aligned}
\end{equation}
where $<|im\_start|>$ and $<|im\_end|>$ are start-of-message and end-of-message tokens of Qwen2-VL.
Specially, we add $<|graph\_start|>$, $<|graph\_pad|>$, and $<|graph\_end|>$ tokens into the vocabulary, where $<|graph\_start|>$ and $<|graph\_end|>$ are start-of-graph and end-of-graph tokens trainable during the last two training stages.
$<|graph\_pad|>$ is a placeholder and its embedding will be replaced by the graph embedding $e_v$ in Equation \ref{eq:pro}.
We show an example of the input instruction in Figure \ref{fig:prompt}

\subsection{Task-oriented Predictor Fine-tuning}
After training the graph feature projector, the LLM can comprehend the node $v$ in the graph $G$.
This stage aims further to improve the ability of the LLM for SPIR.
Since the LLM constructs the attention correlations between the last token and all tokens in $ins$, we leverage the output hidden state feature $f_h$ of the last token. 
We add a task-oriented regression predictor to forecast the influence level, as follows:
\begin{equation}
    \tilde{y}_v=9\cdot sigmoid(\boldsymbol{W}_3f_h+b_3)\in (0,9),
\end{equation}
where $\boldsymbol{W}_3\in\mathbb{R}^{1\times d_{lm}}$ and $b_3\in \mathbb{R}$ are trainable parameters of the predictor.
Finally, we compute the Smooth L1 loss to train the predictor, the LLM, and the graph feature projector jointly, as follows:
\begin{equation}
    \mathcal{L}_{ft} = \frac{1}{|S_{tra}|}\sum_{v\in S_{tra}} SmoothL\textit{1}(\tilde{y}_v, y_v),
\end{equation}
where $S_{tra}$ is the set of all video nodes in the training set and $SmoothL\textit{1}(\cdot,\cdot)$ is the Smooth L1 loss.

%\vspace{-1mm}
\section{Experiments}

%\vspace{-1mm}
\subsection{Experimental Setup}
\textbf{Baseline methods.} We adopt 10 state-of-the-art baseline methods, including 4 GNNs, 4 LLMs, and 2 multimodal LLMs.
The adopted 4 GNN baselines are GCN \cite{kipf2017semi}, HAN \cite{wang2019heterogeneous}, HetSANN \cite{hong2020attention}, and RGCN \cite{zhang2024accessfixer}.
%
%For a fair comparison, we train these GNNs with predictors using the same procedure introduced in Section \ref{sec:hgp}.
%
The adopted 4 LLM baselines are Mistral-7B-Instruct-v0.3 \cite{jiang2023mistral}, InternLM2.5-7B \cite{cai2024internlm2}, Llama-3.1-8B-Instruct \cite{dubey2024llama}, and Qwen2.5-7B-Instruct \cite{yang2024qwen2}.
We further adopt 2 multimodal LLM baselines that can comprehend the video modality, i.e., Llava-Next-Video-7B \cite{li2024llava} and Qwen2-VL-7B-Instruct \cite{wang2024qwen2}.
To conduct a fair comparison, we use the same input graph, predictor module, and optimization loss introduced in Section \ref{sec:hgp} for all GNN baselines.
For LLM and multimodal LLM baselines, we use the same input instructions, output response, and optimization loss introduced in Section \ref{sec:slf}, where the information of the short-video is organized in the json format.
Additionally, we input the video data into multimodal LLM baselines.

\noindent\textbf{Implementation details.}
We adopt a 2-layer RGCN as the graph encoder and Qwen2-VL-7B-Instruct as the LLM backbone in our model.
We utilize the DeepSpeed \cite{rasley2020deepspeed} framework and ZeRO-2 optimizer \cite{rajbhandari2020zero} to train our model on 8 Nvidia A800 80G GPUs.
$d_g$ is set to 1024.
To improve robustness to varying lengths of the input samples while constructing the input instruction, we randomly select 50 comments of a sample if it has more than 50 comments.
Moreover, we set the maximum number of tokens as 4000 when tokeninzing the input instructions.
In task-oriented predictor fine-tuning, we only train the last 4 layers of the LLM decoder while keeping other decoder layers frozen.

For all GNN baselines, we set the dimensionality of hidden and output features as the same as $d_g(=1024)$ in our RGCN encoder.
The number of layers is set to 2 which is also adopted in our GNN encoder.
The minimum sampling number of neighborhood nodes is tuned and set as 50.

For all LLM and multimodal LLM baselines, we apply the same truncation strategy of inputs in our model. We utilize the LoRA 
 algorithm \cite{hu2022lora} to efficiently optimize these models.

\subsection{Metrics} 
\label{sec:met}
Since we annotate the propagation influence of short-videos into the 0 to 9 levels, SPIR can be considered as a 10-class classification task.
However, as the 0 to 9 levels represent a progressive increase in propagation influence, an inaccurate prediction closer to the ground truth is preferable to a prediction that differs significantly.
Therefore, SPIR also exhibits traits of a regression task.
To effectively evaluate the predicted results for SPIR, we adopt both widely-used metrics for classification and regression.
Denote the ground truth for test samples as $\{y_i\in[10]\}_{i=1}^{M}$ and the predicted levels as  $\{\bar{y}_i\}_{i=1}^{M}$, where $y_i$ is the label of the $i$-th test sample, $\bar{y}_i$ is the prediction of the $i$-th test sample, and $M$ is the number of test samples.
We calculate three metrics to evaluate the prediction.

\textbf{Accuracy (ACC).} Following widely-used evaluation for the classification task \cite{haralick1973textural,krizhevsky2012imagenet,qian2023open}, we compute the Accuracy (ACC) of the predicted results as follows:
\begin{equation}
    ACC = \mathrm{P}(y_i=rounding(\bar{y}_i)), i\in [M],
\end{equation}
where $rounding(\cdot)$ is the rounding function.

\textbf{Mean Squared Error (MSE).} Following widely-used evaluation for the regression task \cite{draper1998applied,krizhevsky2012imagenet,tang2022knowledge}, we compute the Mean Squared Error (MSE) of the predicted results as follows:
\begin{equation}
    MSE = \mathrm{E}_{i\in[M]}\{(y_i-\bar{y}_i)^2\}.
\end{equation}

\textbf{Mean Absolute Error (MAE).} Moreover, we compute the Mean Absolute Error (MAE) for regression of the predicted results as follows:
\begin{equation}
    MAE = \mathrm{E}_{i\in[M]}\{|y_i-\bar{y}_i|\}.
\end{equation}

We note that ACC is the higher the better while MSE and MAE are the lower the better.

\begin{table}[t]
\centering
\caption{The results of Short-video Propagation Influence Rating (SPIR) on the XS-Video dataset. $\uparrow$ denotes the higher the better and $\downarrow$ denotes the lower the better.}
%\vspace{-2mm}
\label{tab:res}
\begin{tabular}{c|c|ccc}
\hline
Model            & Input      & ACC$\uparrow$    & MSE$\downarrow$    & MAE$\downarrow$    \\ \hline
GCN              & Graph      & 0.4474 & 1.0623 & 0.7599 \\
HAN              & Graph      & 0.2619 & 2.6666 & 1.2724 \\
HetSANN          & Graph      & 0.5078 & 0.8917 & 0.6803 \\
RGCN             & Graph      & 0.6313 & 0.7801 & 0.5844 \\
Mistral-v0.3     & Text       & 0.5387 & 2.1000 & 0.8123 \\
InternLM2.5      & Text       & 0.5268 & 2.1110 & 0.8064 \\
Llama-3.1        & Text       & 0.5290 & 2.1215 & 0.8081 \\
Qwen2.5          & Text       & 0.5469 & 2.0820 & 0.7688 \\
Llava-Next-Video & Text+Video & 0.5694 & 1.8503 & 0.7315 \\
Qwen2-VL         & Text+Video & 0.5884 & 1.6820 & 0.6629 \\ \hline
\textbf{NetGPT (Ours)}    & \textbf{Graph+Text} & \textbf{0.6777} & \textbf{0.7169} & \textbf{0.5457} \\ \hline
\end{tabular}
%\vspace{-2mm}
\end{table}

\subsection{Results and Discussions}
The experimental results on our XS-Video dataset are shown in Table \ref{tab:res}.
From the results, we have the following observations:
\begin{itemize}[leftmargin=6mm]
    \item 
    %(1) 
    Small GNN models (i.e., GCN, HAN, HetSANN, and RGCN) perform worse than our large graph model (i.e., NetGPT). Compared to the best small GNN (RGCN), our NetGPT relatively increases ACC by 7.3\% and decreases MSE and MAE by 8.1\% and 6.6\%. These results show that bridging graph-structured data with LLM and leveraging large-scale pretrained knowledge can significantly improve SPIR.
    \item 
    %(2) 
    The performance of methods without inputting the propagation graph (i.e., LLMs and multimodal LLMs) is suboptimal compared to RGCN and our NetGPT. Our NetGPT relatively outperforms the best multimodal LLM (Qwen2-VL) by 15.2\%($\uparrow$), 57.4\%($\downarrow$), and 17.7\%($\downarrow$) in terms of ACC, MSE, and MAE, respectively. These results show that capturing the propagation relations in the short-video propagation graph is essential for performing our proposed SPIR task.
    \item 
    %(3) 
    RGCN performs much better than GCN, which further captures the heterogeneity of nodes and edges in the propagation graph. This shows that graph heterogeneity is an important factor in the SPIR task. Moreover, multimodal LLMs with both text and video input outperform LLMs, which shows the content and quality of the videos are also important in SPIR.
\end{itemize}

\begin{table}[t]
\centering
\caption{Ablation results on the XS-Video test set. }
%\vspace{-2mm}
\setlength\tabcolsep{8pt}
% \small
\begin{tabular}{c|ccc}
\hline
Model                  & ACC             & MSE             & MAE             \\ \hline
NetGPT-V             & 0.6564           & 0.7415          & 0.5669           \\
NetGPT-VV              & 0.5889          & 0.8935          & 0.6386          \\
NetGPT-IV              & 0.2805          & 2.2839          & 1.1688          \\
NetGPT-CV              & 0.6464          & 0.8249          & 0.5868          \\
NetGPT-SLF             & 0.6552          & 0.7366          & 0.5605          \\
NetGPT-2B              & 0.6695          & 0.7307          & 0.5563          \\ \hline
\textbf{NetGPT (Ours)} & \textbf{0.6777} & \textbf{0.7169} & \textbf{0.5457} \\ \hline
\end{tabular}
\label{tab:abl}
%\vspace{-2mm}
\end{table}

\begin{figure*}[t]  
    \begin{minipage}{0.32\linewidth}
        \centerline{\includegraphics[width=\linewidth]{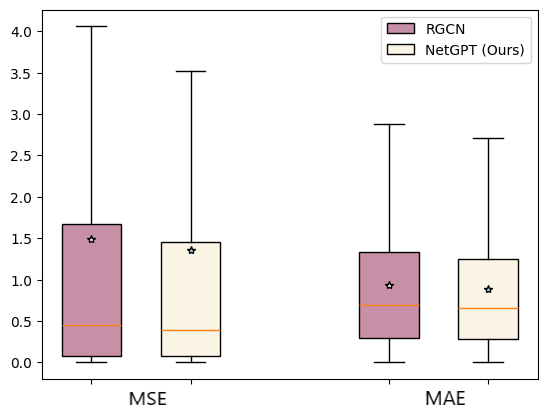}}
        %\vspace{-1mm}
        \centerline{\footnotesize{(a) Long-term prediction}}
    \end{minipage}
    \hfill
    \begin{minipage}{0.32\linewidth}
        \centerline{\includegraphics[width=\linewidth]{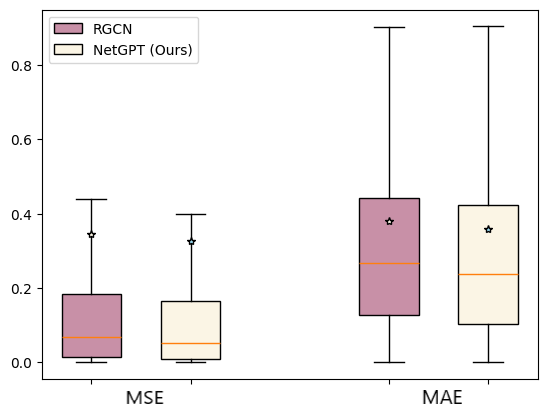}}
        %\vspace{-1mm}
        \centerline{\footnotesize{(b) Median-term prediction}}
    \end{minipage}
    \hfill
    \begin{minipage}{0.32\linewidth}
        \centerline{\includegraphics[width=\linewidth]{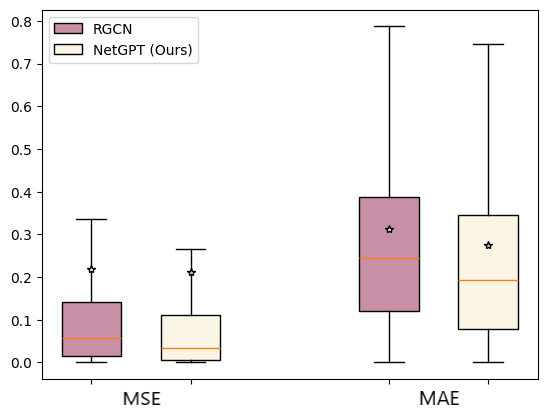}}
        %\vspace{-1mm}
        \centerline{\footnotesize{(c) Short-term prediction}}
    \end{minipage}  
    \vfill
    %\vspace{-3mm}
    \caption{Results of long-, median-, and short-term prediction with the observation times of $\leq 3$ days, $\leq 7$ days, and $>7$ days.}
    %\vspace{-2mm}
    \label{fig:term}
\end{figure*} 

\subsection{Ablation Study}
We design several variants of our NetGPT to conduct the ablation study, as follows: \textbf{NetGPT-V} replaces the raw video features in Equation \ref{eq:vf} by zero vectors; \textbf{NetGPT-VV} deletes all edges among video nodes in the propagation graph; \textbf{NetGPT-IV} deletes all edges from the interactive information nodes (i.e., ctimes, fans, likes, comments, shares, views, collects) to the video nodes; \textbf{NetGPT-CV} deletes all edges from the comment content nodes to the video nodes; \textbf{NetGPT-SLF} abandons the supervised language fine-tuning introduced in Section \ref{sec:slf} in our method; \textbf{NetGPT-2B} replaces the backbone LLM Qwen2-VL-7B-Instruct by Qwen2-VL-2B-Instruct. 

As shown in Table \ref{tab:abl}, we have the following observations: (1) NetGPT-V performs worse than NetGPT, which shows the importance of considering the video content information. (2) NetGPT-VV, NetGPT-IV, and NetGPT-CV perform worse than NetGPT, which shows the edges from video nodes, interactive information nodes, and comment content nodes to video nodes are significant. (3) NetGPT-SLF underperforms NetGPT, which verifies the effectiveness of our three-stage training mechanism. (4) With a smaller LLM backbone, NetGPT-2B performs worse than NetGPT, which indicates the effectiveness of leveraging pretrained knowledge of larger LLMs.

% \subsection{Qualitative Results}
% We demonstrate qualitative results of the predicted answers and explanations for FS-MEVQA in Figure \ref{fig:quali}.
% %
% Compared to VCIN ($N$=16K) and GPT-4V ($N$=16), our MEAgent ($N$=16) can generate more rational and coherent explanations of the reasoning process: 
% %
% (1) In (b)-(d), VCIN cannot ground the key visual objects in the images for explanation, while our MEAgent can utilize the names of objects and an open-world detector to locate accurate boxes. 
% %
% (2) VCIN usually predicts inconsistent answers and explanations. This is because VCIN needs large-scale data to learn the causal correlation between the answer and explanation. Differently, we utilize LLM to translate the execution process of inferring the answer to its explanation, ensuring inherent consistency between answers and explanations.
% %
% (3) In (b)-(d), the detection accuracy of GPT-4V is also unsatisfactory. In (d), GPT-4V grounds an unimportant object (i.e., ``wall") in the explanation but ignores key objects (i.e., ``soap dispensers" and ``bathroom"), which shows GPT-4V fails to capture the correct reasoning process for solving this question. Differently, our MEAgent explicitly generates the multimodal program for solving the input question and generates the explanation accordingly. %Therefore, MEAgent can capture rational reasoning processes more effectively.

\subsection{Results of Short- and Long-term Prediction}
\label{sec:attr}
To further analyze the prediction of different periods, we compute the MSE and MAE of test samples with the observation periods (i.e., current time $-$ post time) of $\leq 3$ days, $\leq 7$ days, and $> 7$ days.
We demonstrate the box diagrams in Figure \ref{fig:term} and compare the results of our NetGPT with the best baseline RGCN.
From the results, we have the following observations: 
(1) For all observation periods, our NetGPT outperforms RGCN in terms of median and average of MSE and MAE. These results further demonstrate the superiority of our method.
(2) For all observation periods, our NetGPT outperforms RGCN in terms of first quartile, third quartile, and maximum of MSE and MAE. These results further demonstrate the better prediction robustness of our method.
(3) For both methods, when the observation period increases, the MSE and MAE of the finally achieved propagation influence levels significantly decrease. These results show the difficulty of our proposed dataset and task that the long-term prediction performance needs improvements.

\begin{figure}[h]  
    \begin{minipage}{0.49\linewidth}
        \centerline{\includegraphics[width=\linewidth]{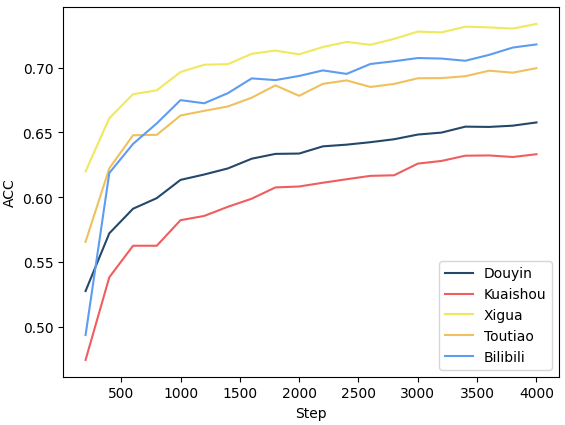}}
        \centerline{\footnotesize{(a) ACC in training procedure}}
    \end{minipage}
    \hfill
    \begin{minipage}{0.49\linewidth}
        \centerline{\includegraphics[width=\linewidth]{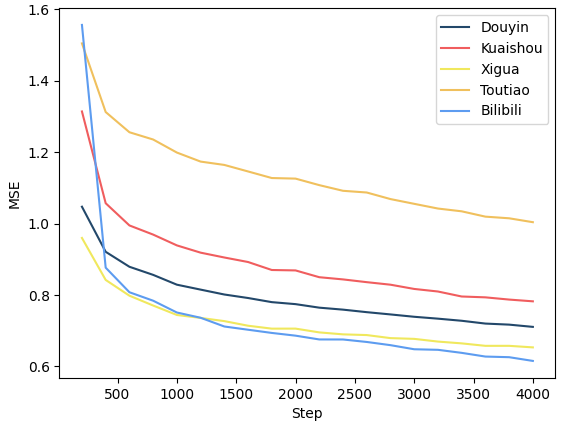}}
        \centerline{\footnotesize{(b) MSE in training procedure}}
    \end{minipage}  
    \vfill
    %\vspace{-2mm}
    \caption{Training procedure on our XS-Video dataset.}
    %\vspace{-2mm}
    \label{fig:proce}
\end{figure} 

%%\vspace{-2mm}
\subsection{Convergence on Different Platforms}
Short-video platforms can differ in sample numbers and propagation patterns.
In this section, we analyze the model convergence on different platforms
In Figure \ref{fig:proce}, we show the ACC and MSE of training samples from different platforms during the 200th-4000th optimization steps in the task-oriented predictor fine-tuning procedure.
(1) The ACC and MSE of Bilibili, which has the fewest samples, quickly converge during fine-tuning, showing the potential over-fitting problem of this small platform.
(3) Interestingly, the ACC and MSE of Xigua, the second largest platform, also converge quickly, which shows that the propagation patterns on Xigua may be relatively easy to recognize.
(3) Overall, 5 platforms exhibit different convergence curves during the training procedure, which indicates the fundamental difference in propagation on these platforms.
Therefore, our proposed XS-Video dataset collects data from these 5 biggest Chinese platforms to facilitate a more comprehensive analysis of short-video propagation.
%%\vspace{-1mm}
\section{Discussion and Conclusion}
In this paper, we propose a Cross-platform Short-Video (XS-Video) dataset with 381,926 samples from the 5 biggest Chinese platforms, including short-video content, post information, interactive information, comment content, and elaborately annotated propagation levels, for short-video propagation analysis. 
Based on our dataset, we propose a new Short-video Propagation Influence Rating (SPIR) task, which aims to predict the long-term propagation influence of a newly posted short-video, under the background of a huge propagation network.
Moreover, utilizing the entities and relations present in XS-Video, we construct a huge propagation graph consisting of 5.5 million nodes and 1.7 billion directed edges.
We build a comprehensive benchmark for SPIR by evaluating state-of-the-art methods, including GNNs, LLMs, and multimodal LLMs. 
To perform SPIR, we propose NetGPT, a novel large graph model to bridge heterogeneous graph data and LLMs.
Extensive experiments demonstrate that NetGPT significantly outperforms state-of-the-art methods for SPIR.

\ifCLASSOPTIONcaptionsoff
  \newpage
\fi

\bibliographystyle{IEEEtran}
\bibliography{sample-base}

\end{document}